\documentclass[10pt,twocolumn,letterpaper]{article}

 \usepackage[pagenumbers]{cvpr} 

\usepackage[dvipsnames]{xcolor}

\usepackage{bbold}
\usepackage{bm}
\usepackage{amsmath}
\usepackage{amssymb}
\usepackage{graphicx}
\usepackage{placeins}
\usepackage{algorithmic, algorithm2e}
\usepackage{subcaption}
\usepackage{natbib}
\usepackage{placeins}
\usepackage{colortbl}

\usepackage{tikz}
\usetikzlibrary{positioning,calc,shapes.geometric,arrows.meta,decorations.pathmorphing,decorations.pathreplacing,calligraphy}
\usetikzlibrary{ decorations.markings}

\newcommand{\suchthat}{\;\ifnum\currentgrouptype=16 \middle\fi|\;}
\newcommand{\indicator}{\mathbb{1}}

\newcommand{\SKIP}[1]{}

\usepackage{pifont}
\newcommand{\cmark}{\ding{51}}%
\newcommand{\xmark}{\ding{55}}%

\newcommand{\dataset}{\mathcal{D}}

\usepackage[dvipsnames]{xcolor}

\definecolor{cvprblue}{rgb}{0.21,0.49,0.74}
\usepackage[pagebackref,breaklinks,colorlinks,citecolor=cvprblue,hypertexnames=false]{hyperref}

\def\paperID{16092} 
\def\confName{CVPR}
\def\confYear{2024}

\title{Attacking Motion Planners Using Adversarial Perception Errors}

\author{Jonathan Sadeghi, 
  Nicholas A. Lord,
  John Redford,
  Romain Mueller \\
  Five AI Ltd. \\
  United Kingdom\\
{\tt\small jonathan.sadeghi@five.ai}
}

\begin{document}
\maketitle

\begin{abstract}

Autonomous driving (AD) systems are often built and tested in a modular fashion, where the performance of different modules is measured using task-specific metrics.
These metrics should be chosen so as to capture the downstream impact of each module and the performance of the system as a whole.
For example, high perception quality should enable prediction and planning to be performed safely.
Even though this is true in general, we show here that it is possible to construct planner inputs that score very highly on various perception quality metrics but still lead to planning failures.
In an analogy to adversarial attacks on image classifiers, we call such inputs \textbf{adversarial perception errors} and show they can be systematically constructed using a simple boundary-attack algorithm.
We demonstrate the effectiveness of this algorithm by finding attacks for two different black-box planners in several urban and highway driving scenarios using the CARLA simulator.
Finally, we analyse the properties of these attacks and show that they are isolated in the input space of the planner, and discuss their implications for AD system deployment and testing.

\end{abstract}

\section{Introduction}
\label{sec:introduction}

In safety-critical systems such as autonomous driving, it is crucial to establish as much as possible about real-world performance prior to real-world deployment.
High-severity, low-probability failures are especially important to capture and characterise as these are the ones most likely to be missed during standard development and testing~\cite{koopman2016challenges}.

Current testing methodologies consist of a careful elucidation of the operational design domain (ODD) in which the system will be deployed, and specification of the desired behaviour of the system in the ODD via the definition of driving rules \cite{koopman2019many, riedmaier2020survey, koopman2019safety}.
This allows the behaviour of the system as a whole to be assessed on the basis of how often driving rules are broken, which is essential for safe deployment.
Individual components making up the system can also be tested separately: however, the performance of the perception module when tested with common metrics like mean average precision might only be weakly correlated with the impact of perception mistakes on the planning system \cite{philion2020learning}.
The system as a whole as well as the individual subsystems should be fine-tuned on recorded data and in simulation prior to deployment.
On deployment of the system in the real world, further data can be collected which can be used to improve the system in the future \cite{koopman2018toward}.

In this work we demonstrate the existence of sets of erroneous perception system outputs which score highly in common perception metrics, but nevertheless cause the system to break driving rules.
We therefore term these sets of perception errors \emph{adversarial perception errors}.
The existence of these adversarial perception errors has implications for how these systems are built and tested and is therefore highly relevant to practitioners in the field of autonomous vehicles.
Leveraging ideas from adversarial attacks on image classifiers \cite{brendel2018decisionbased}, we provide an efficient search algorithm which yields the most adversarial \emph{perception failure modes} for the system in simulation, where the importance of these modes is assessed by the user-specified perception metric.
We test our algorithm in the CARLA simulator \cite{Dosovitskiy17} on a recent optimisation-based planner \cite{eiras2021two} and a lane-keeping planner based on the Intelligent Driver Model \cite{treiber2000congested}.
We judge the importance of the identified errors using the nuScenes detection score and other metrics, and analyse the wider impact of our findings for autonomous vehicle development.
\section{Background}
\label{sec:background}

At any given time $t$, the agents in a driving environment can be described by a state $s_t \in \mathcal{S}$, which contains the properties of every agent in the scene (\eg position, velocity, etc.) as well as sensor data like LiDAR point clouds and RGB images.
Given state $s_t$ at time $t$, let us assume that the system takes an action $a_t \in \mathcal{A}$ and define a $T$-step rollout as $\tau = \left[ s_0, a_1, s_1, a_2 \dots s_{T - 1} \right]$ with $s_t \sim p(s_t \mid s_{t-1}, a_{t})$.
We assume here that the behaviour of other agents in the scene is deterministic and that $s_{t}$ can be determined completely given state $s_{t-1}$ and action $a_t$, i.e.~$p(s_t \mid s_{t-1}, a_t)$ is a delta function. 
In many cases, non-deterministic agent actions can be made deterministic by parameterising the agent behaviour in some way, \eg by specifying the aggressiveness and direction of turns by an agent in a particular scenario, and therefore we do not regard this assumption as overly restrictive.
We further assume that the simulator can be made deterministic, see \cite{chance2022determinism}.

We consider driving agents that rely on a perception system to build a representation of the world and use this representation to plan and act.
This does not apply to end-to-end driving systems which we do not consider here \cite{bewley2019learning, tampuu2020survey}.
We represent the perception system as a function $f: \mathcal S \to \mathcal{\hat{S}}$ that maps an environmental state $s$ to a perceived state $\hat{s} = f(s)$ deterministically.
This could be, for example, a camera- or lidar-based 3D object detector.
Let us further assume that the system plans and acts deterministically given the perceived state and denote its policy by $\pi$.
This means that at time $t$ the action $a_t$ is chosen as
\begin{equation}
    a_t = \pi(\hat s_t) = \pi(f(s_t)).
\end{equation}
The set of perceived states $\mathcal{\hat{S}}$ is in general different from the set of states $\mathcal{S}$, \eg~the number of perceived agents can be different, and the agents might be parameterised differently.

\paragraph{Perception quality}
\label{sec:perception_metrics}

The quality of the perception system can be assessed using a set of task-specific metrics that characterise the deviation of a perceived state $\hat s = f(s)$ from the corresponding environmental state $s$.
These include, for example, mean average precision or the nuScenes detection score~\cite{caesar2020nuScenes}.
More formally, for any sequence of ground-truth and perceived scenes $y = \left[ s_0,  s_1, \dots, s_{T - 1} \right]$ and $\hat{y} = \left[ \hat{s}_0, \hat{s}_1, \dots, \hat{s}_{T - 1} \right]$, we can define a perception metric as a real valued function $m(\hat{y}, y) > 0$ which measures the quality of the perception for the entire sequence.
We assume that higher perception scores indicate better perception and that $m(y, y) = 1$.

\paragraph{Driving rules}

The performance of the overall driving system can be tested against a set of driving rules that encompass both safety and other aspects of driving such as comfort.
We consider here rules with binary pass/fail outcomes, where failure indicates behaviour that is unacceptable for the driving system (e.g~a collision).
For a given a rollout $\tau$, we implement driving rules using real-valued functions $r(\tau) \in \mathbb{R}$ such that the condition $r(\tau) < 0$ denotes violation of the corresponding rule.
For example, the metric corresponding to a collision could be the closest distance of approach of ego to any other agent.
The performance and safety of the system can then be assessed by computing the average rate of failures over a specified number of scenarios (also known as probabilistic threshold robustness \cite{beyer2007robust}).

\paragraph{Link between perception and driving performance} 
The link between perception quality and overall system performance is generally complex.
Even though it is expected that better perception will make it easier for the system as a whole to drive safely, the exact relationship between module-level and overall driving performance is unclear.
In what follows, we show that it is possible to find perception errors $\hat y$ that score highly with respect to the perception quality metrics ($m(\hat y, y) \approx 1$) but that still lead to the planner violating the driving rules ($r(\tau) < 0$).

\begin{figure*}[ht!]
    \centering
    \begin{tikzpicture}[node distance=0.8cm,scale=0.65]
\tikzset{every node/.style={transform shape, align=center, draw, rounded corners=3, inner sep=2mm, thick}}

\node[fill=cyan!20] (simulator) {Simulator};
\coordinate[right=of simulator] (c1);
\node[right=1.7cm of c1, circle, inner sep=1mm, fill=pink!80] (attack) {$I$};
\coordinate[right=2.3cm of attack] (c2);
\node[right=of c2, fill=cyan!20] (tracker) {Tracker};
\node[right=of tracker, fill=cyan!20] (planner) {Planner};
\node[above=0.8cm of attack, fill=green!20] (metric) {Perception\\Metric $m(\hat y, y)$};
\node[draw=none, below=0.3cm of attack, inner sep=1mm] (errors) {Perception errors $\mathbf e$};
\node[below=0.3 of errors, fill=pink!80] (search) {Search\\Algorithm};
\node[fill=green!20] at (search-|planner) (eval) {Driving Rule\\$r(\tau) \lessgtr 0$};
\coordinate[above=0.6cm of metric] (top);

\draw[thick] (simulator) -- (c1);
\draw[->, thick] (c1) -- (attack);
\draw[thick] (attack) -- (c2);
\draw[->, thick] (c2) -- (tracker);
\draw[->, thick] (tracker) -- (planner);
\draw[->, thick] (planner.north) |- (top);
\draw[thick] (simulator.north) |- (top);
\draw[dashed] (search) -- (errors);
\draw[->, dashed] (errors) -- (attack);
\draw[->, dashed] (c1) |- (metric);
\draw[->, dashed] (c2) |- (metric);
\draw[->, dashed] (planner) -- (eval);
\draw[->, dashed] (eval) -- (search);

\node[draw=none, above=0.4cm of c1, fill=white, inner sep=1mm] {World\\State $y$};
\node[draw=none, above=0.4cm of c2, fill=white, inner sep=1mm] {Perceived scene\\$\hat y = I(y, \mathbf e)$};

\end{tikzpicture}
    \hspace{2cm}
    \input{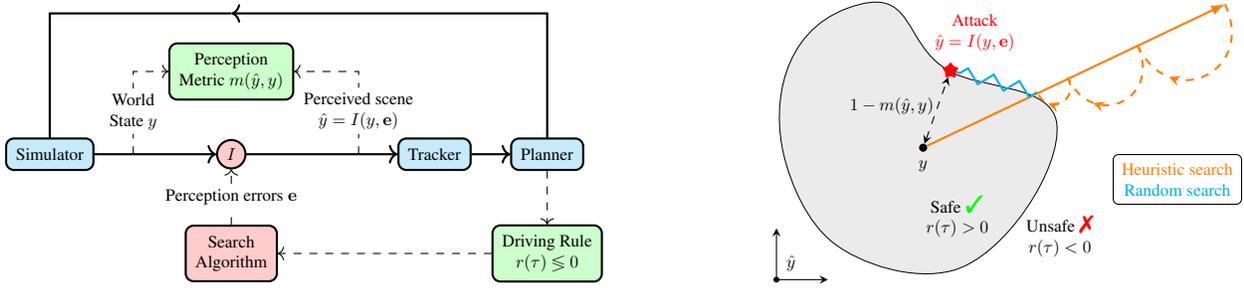}
    \caption{
    Adversarial perception error search.
    Left:
    Starting from a standard simulation rollout without perception system (blue), we inject perception errors $e$ to create the the perceived scene $\hat y = I(y, \mathbf e)$ (red) and search for perception errors $\mathbf e$ that make the planner fail while maximising the perception metric $m(\hat y, y)$ (green), see text for details.
    Note that dashed lines represent actions that occur once after every completed rollout.
    Right:
    A detailed graphical representation of our search strategy showing the heuristic (orange) and random (cyan) searches.
    }
    \label{fig:approach}
\end{figure*}

\section{Approach}
\label{sec:methodology}

We propose a simple method for identifying adversarial perception errors which is applicable for black-box systems, i.e.\ those for which gradients are not available.
As discussed above, we regard perception errors as adversarial if they result in rule-breaking behaviour whilst also having high perception quality.
Given a perception metric $m$, we adopt a fuzzy set construction and define the set of perception errors that have perception quality of at least $\alpha$ as $\mathcal{Y}_{\alpha}=\{ \hat{y} \mid m(\hat{y}, y) > \alpha \}$, where $y$ and $\hat y$ are the real and perceived scene.
Following the notation of \cref{sec:background}, we define the set of rollouts with perception quality of at least $\alpha$ as
\begin{align}
    T(\alpha) &= \left\{\tau = \left[ s_0, a_1, s_1, a_2, \dots, s_{T - 1} \right] \mid \right. \nonumber\\ 
    & \qquad \hat{y} = [\hat s_0, \hat s_1, \ldots, \hat s_{T-1}] \in \mathcal{Y}_{\alpha},\nonumber\\
    & \qquad a_t = \pi(\hat{s}_{t-1}), \left. s_t \sim p(s_t \mid s_{t-1}, a_{t})\right\}.
\end{align}
The task of finding adversarial attacks can then be formulated as finding the largest $\alpha$ such that there is at least one rollout in $T(\alpha)$ failing the driving rule $r$, i.e.~we want to
\begin{equation}
    \text{maximise $\alpha$ such that $\min_{\tau\in T(\alpha)}r(\tau)\hspace{0.3em} <0$}. \label{eqn:set-sim}
\end{equation}
Any perceived scene sequence $\hat y$ with maximal $\alpha$ for which $r(\tau)<0$ is to be considered an adversarial attack.

Solving \cref{eqn:set-sim} exactly is very difficult as it requires checking all possible perceived states $\hat y$.
Instead, we opt for finding increasing lower bounds for the maximum perception quality $\alpha$ by searching for explicit examples of failing rollouts with increasing perception quality $\alpha$.
To do so, we parameterise the perceived scene sequence $\hat{y}$ explicitly as
\begin{align}
    \hat{y} &= \left[ \hat{s}_0, \hat{s}_1, \dots, \hat{s}_{T - 1} \right] \\
    &= \left[ I({s}_0, e_0), I({s}_1, e_1), \dots, I({s}_{T-1}, e_{T-1}) \right] = I(y, \mathbf e),\nonumber
\end{align}
where $y = [s_0, \ldots, s_{T-1}]$ is the ground-truth state of the world and $I$ is a parametric attack function with perception error parameters $\mathbf e = [e_0, \ldots, e_{t-1}]$.
We can then obtain the corresponding rollout $\tau$ and check for violation of the driving rules $r(\tau) < 0$.
This approach is illustrated on \cref{fig:approach}.
We use $e_i = [(\mathbf{x}_1, \phi_1, \text{fn}_1), \ldots, (\mathbf{x}_d, \phi_d, \text{fn}_d)]$, where $d$ is the number of agents in the scene, $\mathbf{x}_j$ is a Cartesian-additive error for agent $j$, $\phi_j$ is an orientation-additive error for agent $j$, and $\text{fn}_j$ is a binary ``false-negative'' switch that causes agent $j$ to be completely removed from the output of $I$.
Of course, many other parameterisations are possible.

We split out algorithm in two phases: a heuristic and a random search.
The heuristic search is a hand-crafted strategy that aims at finding a perceived scene sequence close to the failure boundary surrounding the ground-truth $y = I(y, \mathbf e)$ as quickly as possible.
A random search is then applied to refine the attack further by increasing $\alpha = m(\hat y, y)$ using random steps while keeping $\hat y$ in the failure region, see \cref{fig:approach} left for an illustration.
Directly applying random search around the ground-truth $y$ would fail to lead any improvements because we expect the system to be resilient to small errors $\mathbf e$, so most steps would be rejected due to not finding any rule violations.
This approach is inspired by the Boundary Attack algorithm to find adversarial attacks on black-box models in the image space~\cite{brendel2018decisionbased}.

\paragraph{Heuristic Search}
The heuristic search algorithm is designed to efficiently find rollouts such that $r(\tau) \approx 0$ using a simple bisection approach.
Our algorithm is based on the intuition that if the perception system would detect no agents at all, then a driving rule violation is very likely to occur, and that detecting more agents more of the time would most certainly improve the perception metric.
We first find the influential agents in the scene by performing a different rollout for each agent where the entire track for the agent is not perceived (i.e. a 100\% false negative error), which corresponds to errors of the form
\begin{align}
    \mathbf{e}^{(j)}& = [e^{(j)}_{0}, e^{(j)}_{1}, ..., e^{(j)}_{(T-1)}] \nonumber \label{eqn:heuristic} \\
 e^{(j)}_{t} &= [(\mathbf{x}_i=\mathbf{0}, \phi_i=0, \text{fn}_i = \delta_{ij}) : i=1, \ldots, d], 
\end{align}
and select those agents that lead to a collision when dropped.
Then, for each  influential agent, we find the minimum track drop time required to cause a collision by running a bisection algorithm both for the start and end times of the false-negative part of the track.
This can be achieved by writing an error sequence where only the time segment between $t_1$ and $t_2$ is dropped i.e. $\mathbf{e}^{(j)}_{(t_1, t_2)} = [e_\emptyset ~\text{for}~ t \in [t_1, t_2],  e^{(j)}_{t}~\text{otherwise}], \label{eqn:bisection}$
where
$ e_\emptyset = \{(\mathbf{x}_i=\mathbf{0}, \phi_i=0, \text{fn}_i = 0) : i =1, \ldots, d\}$.
This approach is described more formally in \cref{alg:1}.

\paragraph{Random Search}
As detailed in \cref{alg:2}, we sample small random steps in $\mathbf e$ using the proposal distribution described below and accept the step if the resulting perceived state $\hat y = I(y, \mathbf e)$ has higher $\alpha = m(\hat{y}, y)$ and still leads to a planning failure, $r(\tau) < 0$.
We use a proposal distribution conditional on the previous error $p(\mathbf{e}|\mathbf{e}^{i-1})$ which is biased towards increasing $\alpha$: 
we replace false negatives from the original heuristic search with true positives for a random segment length of the false negative part of the track and add some small random spatial and orientation noise to the new true positive detections. 
Of course, other proposal distributions are possible.
To decrease the number of simulations, we reject random steps if they lower the perception metric using the previous step's rollout and perform a full rollout only for accepted steps.

\RestyleAlgo{ruled}
\begin{algorithm}
    \caption{Heuristic Perception Error Search}
    \label{alg:1}
    \begin{algorithmic}
        \REQUIRE Rule $r$, simulator for rollout ($\tau$) generation.
        \STATE Obtain the set of times and agents in the simulation when running with ground truth, i.e. obtain $\tau$ for $\hat{s}$ directly corresponding to $s$.
        \STATE $d$ = number of agents
        \FOR{j = 1,\ldots, $d$}
        \STATE Obtain rollout $\tau$ with $\mathbf e_j = \mathbf e^{(j)}$ from \cref{eqn:heuristic}
        \IF{$r(\tau) < 0$}
        \STATE Find largest $t_\text{start}$ such that $r(\tau) < 0$ using bisection by running rollouts with $\mathbf{e}^{(j)}_{(t_\text{start}, T-1)}$.
        \STATE Find smallest $t_\text{end}$ such that $r(\tau) < 0$ using bisection by running rollouts with $\mathbf{e}^{(j)}_{(t_\text{start}, t_\text{end})}$.
        \STATE Set $\mathbf e_j = \mathbf{e}^{(j)}_{(t_\text{start}, t_\text{end})}$
        \ENDIF
        \ENDFOR
        \ENSURE Failure modes $ \mathbf{e}_j, j=1, \ldots, d$
    \end{algorithmic}
\end{algorithm}

\RestyleAlgo{ruled}
\begin{algorithm}
    \caption{Perception Error Random Search}
    \label{alg:2}
    \begin{algorithmic}
        \REQUIRE Rule $r$, simulator for rollout ($\tau$) generation, perception metric $m$, parametric attack function $I$, random error step generator $p(\mathbf{e}|\mathbf{e}^{i-1})$, initial error $\mathbf{e}^0$.
        \STATE Run simulation with $\mathbf{e}^0$ to obtain $\tau$ and $y^0$ 
        \STATE $\alpha_0 = m(\hat{y}, y^{0})$, $\hat{y} = I(y^{0}, \mathbf{e}^0)$
        \FOR{i = 1,\ldots, $N_\text{steps}$}
        \FOR{j = 1,\ldots, $N_\text{proposal-steps}$}
        \STATE Sample $\mathbf{e}^j_{\text{proposal}} \sim p(\mathbf{e}|\mathbf{e}^{i-1})$
        \STATE $\hat{y}^j = I(y^{i-1}, \mathbf{e}^j_{\text{proposal}})$ 
        \ENDFOR
        \IF{$\text{any}_j(m(\hat{y}^j, y^{i-1})) > \alpha_{i-1}$}
        \STATE Set $\hat{y}$ and $\mathbf{e}$ as a randomly chosen $\hat{y}^j$ and $\mathbf{e}^j_{\text{proposal}}$ such that $m(\hat{y}, y^{i-1}) > \alpha_{i-1}$
        \STATE Run simulation with $\mathbf{e}$ to obtain $\tau$ and $y^i$ 
        \ENDIF
        \IF{$\text{any}_j(m(\hat{y}^j, y^{i-1})) > \alpha_{i-1}$ and $r(\tau) < 0$} 
                \STATE $\alpha_i = m(\hat{y}, y^i)$, $\mathbf{e}^i = \mathbf{e}$
        \ELSE
        \STATE  $\alpha_i = \alpha_{i-1}$, $\mathbf{e}^i=\mathbf{e}^{i-1}$
        \ENDIF
        \ENDFOR
        \ENSURE $\alpha_{N_\text{steps}}$ (approximate solution to \cref{eqn:set-sim}) and $\hat{y}$
    \end{algorithmic}
\end{algorithm}

\section{Related Work}
\label{sec:related}
Here we provide a brief summary of related work; an extended related work section is given in \cref{sec:related_full}.

\paragraph{Adversarial Attacks}
Inputs of RGB-image-based deep learning systems which appear benign to humans but cause unexpected behaviour, such as the system predicting incorrect classes, are often described as \emph{adversarial} \cite{szegedy2013intriguing}.
Such attacks may be performed in the real world \cite{ wu2020making, lee2019physical, li2019adversarial, van2019fooling, sharif2016accessorize, chen2019shapeshifter, kurakin2018adversarial,brown2017adversarial,eykholt2018robust,song2018physical}, including on autonomous vehicles \cite{morgulis2019fooling,zhang2018camou,sitawarin2018darts}.
In the `decision based' setting, where only the predicted class of the classifier is known, an algorithm was proposed by \citet{brendel2018decisionbased} to identify these failure modes. 
Furthermore, more efficient iterations of this algorithm have been developed \cite{Dong_2019_CVPR, 9152788, cheng2019query, sitawarin2022preprocessors}.

\paragraph{Adversarial Scenarios}
The identification of useful and representative driving scenarios which can be used to effectively test autonomous vehicles, whilst not necessarily appearing benign to humans, has emerged as a separate task from the overall estimation of failure probability for the system \cite{corso, zhang2021finding}.
Many techniques have been used to search for these scenarios and make the search computationally feasible; for example, surrogate model techniques \cite{sinha2020neural, beglerovic2017testing, vemprala2021adversarial}, reinforcement learning \cite{corso2019adaptive, koren2018adaptive, koren2019efficient}, and approximate gradients from differentiable physics models \cite{hanselmann2022king}.

\paragraph{Reliability Analysis}
Many simulation techniques used to test AD systems were invented prior to the advent of AD.
For example, approximations of the system performance can be used to determine the system's most likely failure mode (the \emph{design point}), which in turn determines the system's failure probability \cite{hasofer1974exact, rackwitz1978structural, hohenbichler1984asymptotic, Breitung1994}.
The reliability of a system can be evaluated by modelling uncertain system variables as fuzzy sets \cite{moller2004fuzzy, moller2000fuzzy}, which is similar to the use of a metric to specify a level of performance for the perception system used in this work.
Sometimes the associated probability of the failure modes is calculated using efficient sampling techniques \cite{uesato2018rigorous}, or surrogate models \cite{inatsu2021active, sadeghi2022bayesian}.

\section{Experiments}
\label{sec:experiments}

In this section we show that the simple boundary-attack algorithm presented above is able to to systematically construct adversarial perception errors in a variety of scenarios.
We consider a simple system configuration consisting of a 3D object detector, a simple object tracker, and a planner.
We attack two different black-box planners in five different urban and highway driving scenarios, and finally discuss the implications of our results for AD system deployment and testing.

\subsection{System Setup}
\label{sec:experiments:setup}

\paragraph{Object detector and perception metrics}
We use the BEVFusion 3D object detector~\cite{liu2022bevfusion} which is a state-of-the-art camera-lidar fusion detector on the nuScenes dataset.
It outputs bounding boxes with pose, size, and velocity in top-down space.
We use the default settings and weights provided in the original implementation.

\begin{figure}
    \centering
    \includegraphics[scale=1]{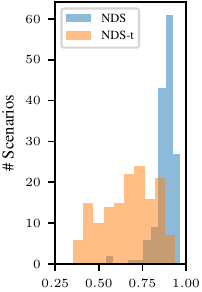}
    \includegraphics[scale=1]{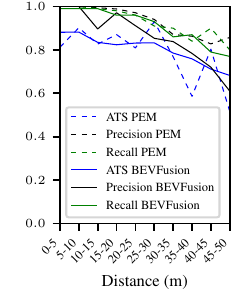}
    \caption{Performance of BEVFusion~\cite{liu2022bevfusion} and comparison with our perception error model (PEM) on the nuScenes dataset.
    (left) Distribution of NDS and NDS-t scores achieved by BEVFusion on the nuScenes val dataset.
    (right) Comparison of BEVFusion and our PEM on nuScenes test set.
    ATS (average translation score) is defined as one minus the average translation error for true positives.
    We can see that our PEM reproduces the errors from BEVFusion reasonably well.
    }
    \label{fig:pem}
\end{figure}

We measure the performance of the object detector using the following scene-level perception quality metrics:
\begin{itemize}
 \item \emph{nuScenes detection score (NDS)}: a weighted combination of mean average precision and various box-level errors (translation error, orientation error, etc.)  \cite{caesar2020nuScenes}.
    \item \emph{NDS with continuous false negative penalty (NDS-t)}: equally weighted sum of the NDS and a term penalising the longest fraction of the track which is a continuous false negative.
\end{itemize}
The NDS-t metric considers flickering detections less critical because they are usually removed by the tracker, which means that contiguous false negatives have a more severe impact on planning performance.
This is not captured in the original NDS.
Equations are given in \cref{sec:metrics}.
In \cref{fig:pem} (left) we show a histogram of perception metric functions on the nuScenes val dataset for the BEVFusion detector.
In all perception metric functions we remove any object categories other than cars, since cars are the only category of actor used in this study.

\paragraph{Tracker} We track detections from the object detector using a simple Kalman-filter-based multi-object tracker inspired by~\cite{bewley2016simple}.
We use the location and orientation of the boxes in the 2D BEV space as state variables.
We use a constant velocity model for the position and the orientation, which we encode as $(\cos(\theta), \sin(\theta))$ and re-normalise at each time step.
We associate detections to tracks using the Hungarian algorithm using the distance between box centres as the cost matrix with a threshold of 2 metres \cite{forsyth2012computer}.
Tracks are confirmed after one observation, and are deleted if unobserved for 1 second, aligning with the planning decision interval.
We only consider the false negative, orientation, and spatial position error properties for the observed states --- see \cref{sec:methodology} --- and therefore give the tracker access to the ground-truth values of the other observed variables.

\begin{figure}
    \centering
    \hfill
    \begin{subfigure}[b]{0.12\columnwidth}
    \centering
    \includegraphics[scale=0.15]{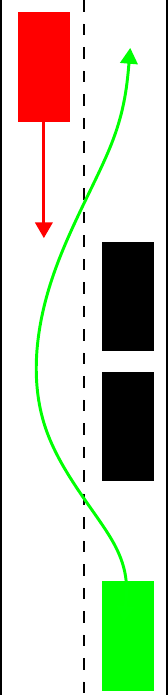}
    \caption{}
    \end{subfigure}
    \hfill
    \begin{subfigure}[b]{0.25\columnwidth}
    \includegraphics[scale=0.15]{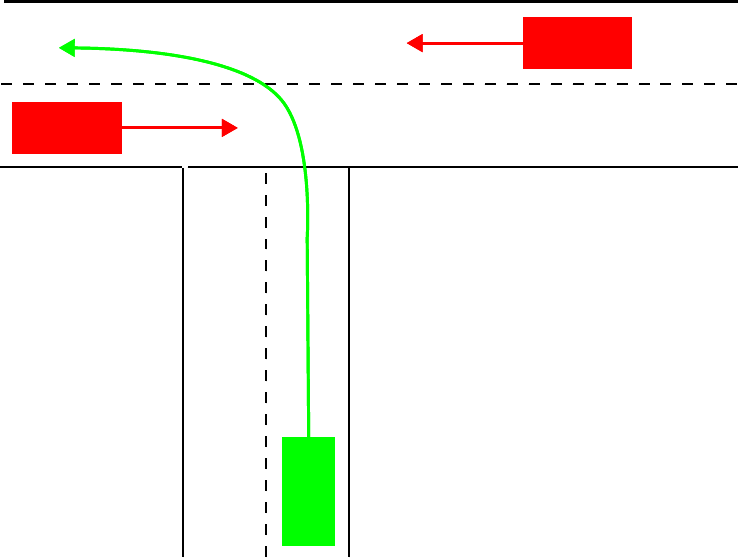}
    \caption{}
    \end{subfigure}
    \hfill
    \begin{subfigure}[b]{0.25\columnwidth}
    \includegraphics[scale=0.15]{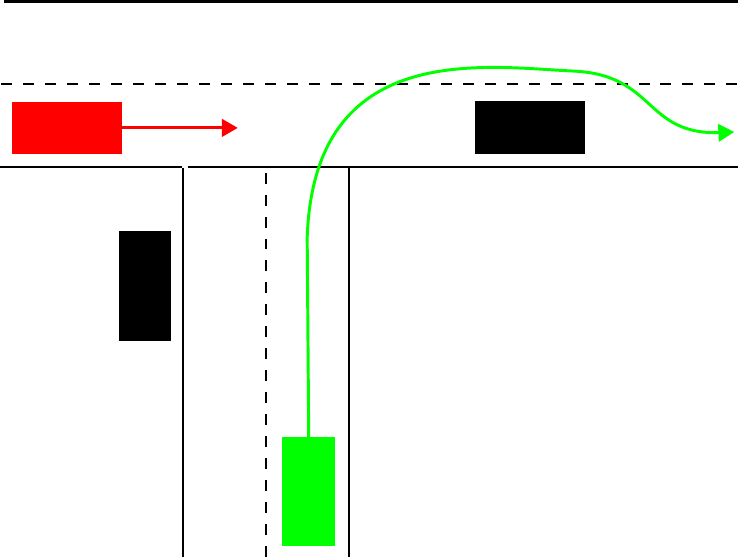}
    \caption{}
    \end{subfigure}
    \hfill
    \begin{subfigure}[b]{0.12\columnwidth}
    \centering
    \includegraphics[scale=0.15]{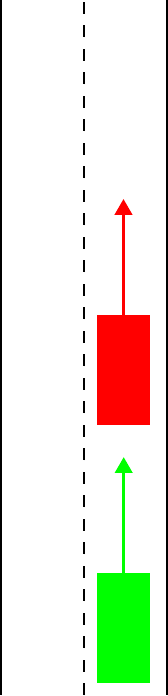}
    \caption{}
    \end{subfigure}
    \begin{subfigure}[b]{0.12\columnwidth}
    \centering
    \includegraphics[scale=0.15]{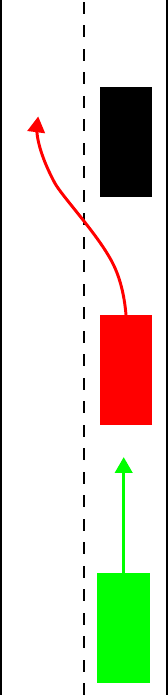}
    \caption{}
    \end{subfigure}
    \hfill
    \caption{Test scenarios for the ObP: (a) overtake, (b) left turn, (c) right turn; and IDM planners: (d) lane following, (e) overtake follow.
    The diagrams show the road configuration and the positions and trajectories of the different vehicles (green=ego, red=moving, black=stationary).}
    \label{fig:scenario-diagrams}
\end{figure}

\paragraph{Planners and tasks}
We test the following two planners on a selection of tasks within their operational design domain depicted in \cref{fig:scenario-diagrams}:
\begin{itemize}
    \item \emph{ObP}: An optimisation-based planner \cite{eiras2021two}.
    \begin{enumerate}
        \item Overtaking a stationary vehicle with a vehicle moving at constant velocity in the oncoming lane
        \item Turning left at a T-junction, into the far-side lane across oncoming traffic travelling at constant velocity
        \item A T-junction right turn, into the nearside lane while avoiding traffic
    \end{enumerate}
    \item \emph{IDM}: A path-following planner based on the Intelligent Driver Model \cite{treiber2000congested, dauner2023parting}.
    \begin{enumerate}
        \item Lane-following a constant-velocity vehicle
        \item Lane-following a vehicle which overtakes a stationary vehicle
    \end{enumerate}
\end{itemize}

\paragraph{Hyperparameter tuning}
In order to make our setup as realistic as possible, we tune its hyperparameters using a simple Perception Error Model (PEM) trained on the output of the BEVFusion detector on the nuScenes dataset.
A PEM is a probabilistic model of the distribution of the perceived objects conditioned on the ground-truth state of the world~\cite{piazzoni2020modeling, piazzonimodeling, zec2018statistical, mitra2018towards, sadeghi2021step, innes2023testing} and allows us to model the error behaviour of BEVFusion on nuScenes, while running scenarios in the CARLA simulator.
Specifically, we follow~\cite{sadeghi2021step} and train a lightweight feedforward network to predict the existence of true positive detections and their spatial errors, see \cref{sec:pem} for details. 
Ground-truth data is obtained by running the BEVFusion detector on the nuScenes dataset and then associating the ground-truth objects to detections to obtain lists of true positive and false positive detections.
We split the nuScenes validation set sequentially, with the first 90\% of scenes used to train the model and the final 10\% of scenes used for testing.
\cref{fig:pem} shows the test set performance of the PEM --- further analysis is shown in \cref{sec:pem_nds}.
Using this PEM, we make our system robust to \emph{typical} perception errors by choosing hyperparameters such that no collisions were observed when taking 100 random samples from the PEM in each scenario.
This gives us some confidence that our system is robust to BEVFusion's typical errors on nuScenes.

\paragraph{Simulator}
We use the CARLA simulator and perform simulation rollouts from \cref{alg:1} and \cref{alg:2} using an ``open-loop'' simulation approach, where we only compute the state sequences of the other agents $s$ once without perception errors (in the world frame), and then apply the attack function to these ground-truth states to obtain the perceived state, $\hat{s}$, which can be used to recompute the ego plans.
This approach is much more efficient because it avoids the expense of repeatedly performing rollouts with the full simulator to obtain $\tau$, and also avoids simulator non-determinism.
Although the state sequence of other agents will be frozen in the world frame, it will change in the ego-centric frame since the plans of ego will depend on the applied perception errors during the attack.
Because the tasks presented above do not present much interaction between the ego and the other agents, we have seen very little difference between this approach compared to a full ``closed-loop'' simulation and use the former in the rest of this paper.

\begin{figure*}

\begin{minipage}{.4\textwidth}
   \includegraphics[trim={0 0.03cm 0 0.03cm},clip]{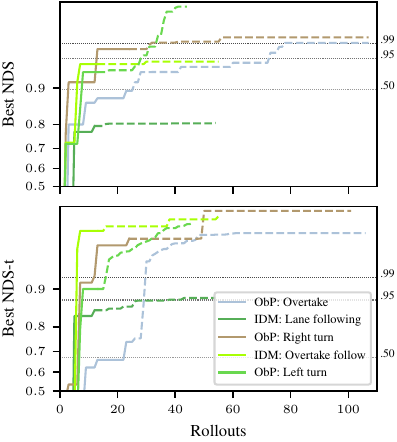}    
   \end{minipage}
\begin{minipage}{.5\textwidth}
    \begin{tabular}{clll>{\columncolor{red!10}}c}
        \hline
        &&&& \\
        \rotatebox{90}{\hspace{0.5cm}\scriptsize ObP: left turn} &
        \includegraphics[width=0.23\columnwidth]{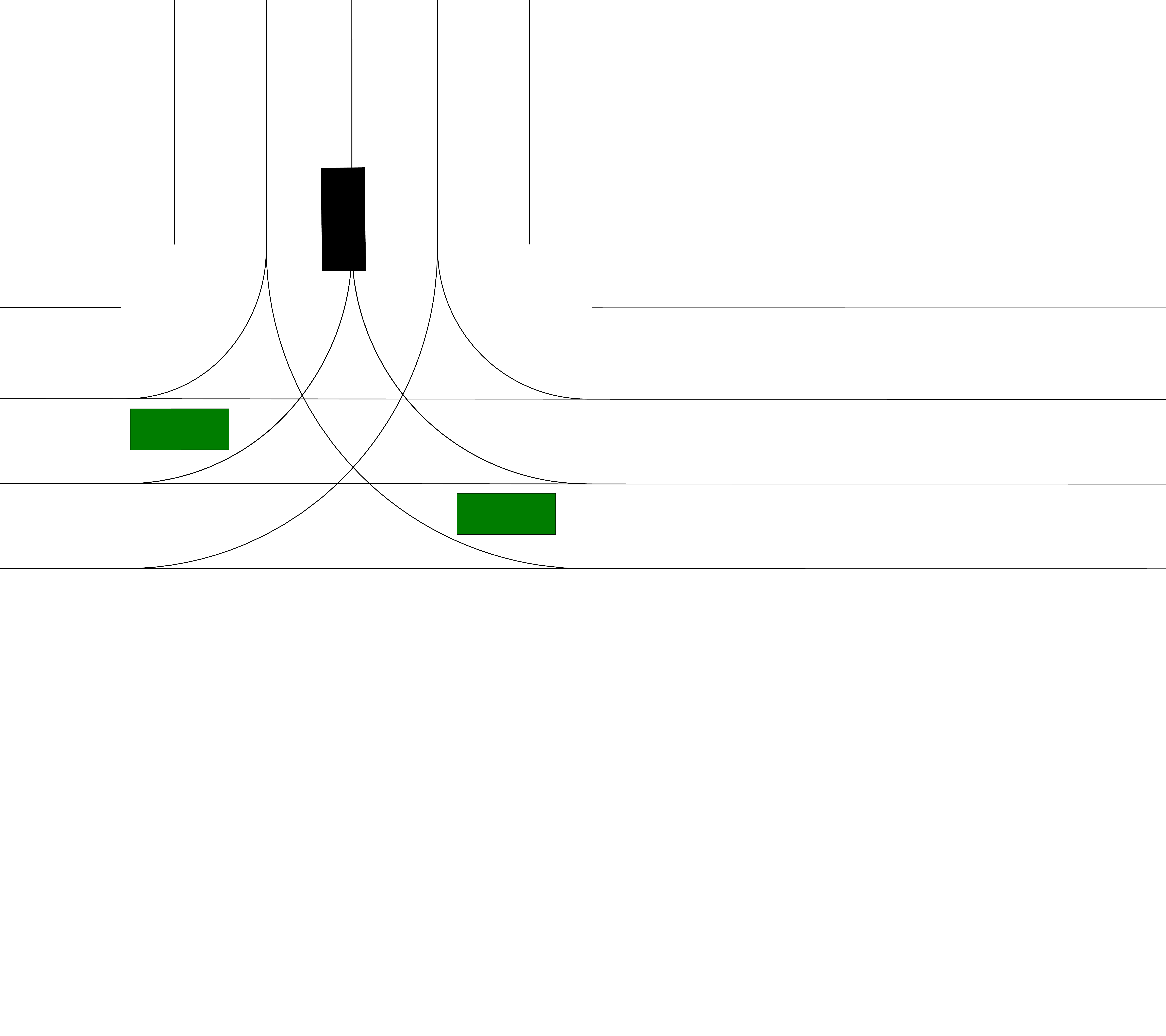} &
        \includegraphics[width=0.23\columnwidth]{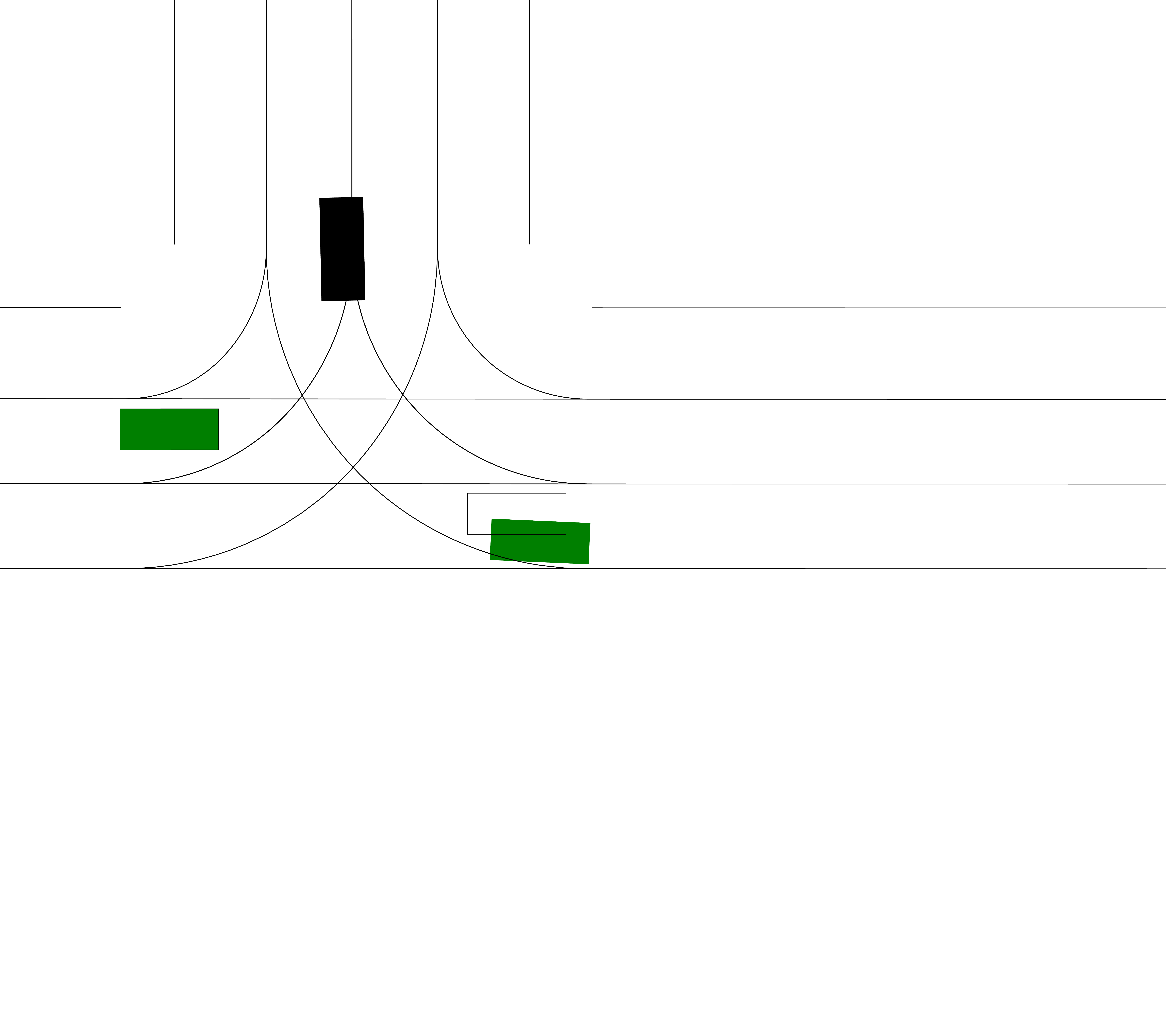} &
        \includegraphics[width=0.23\columnwidth]{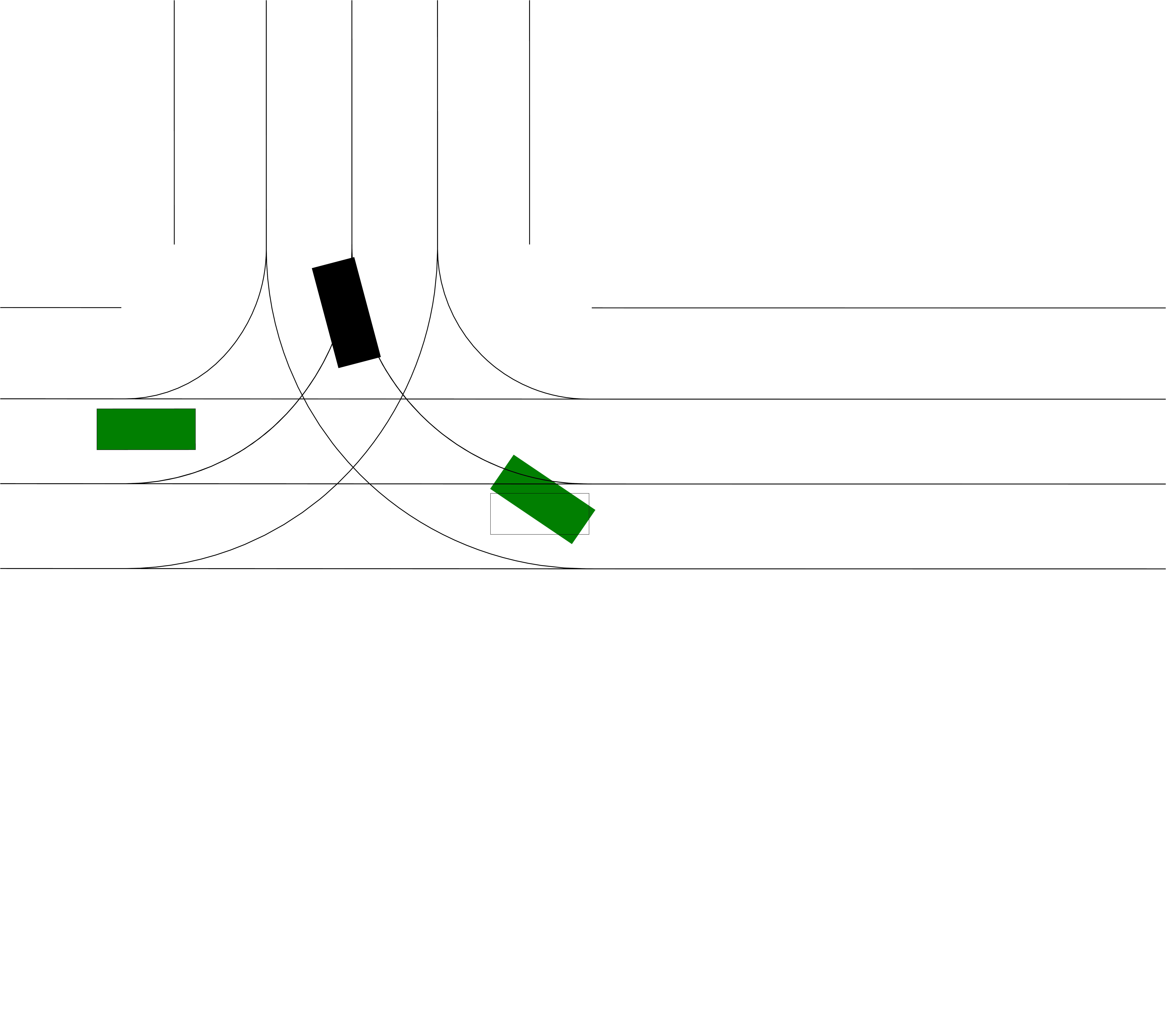} &
        \includegraphics[width=0.23\columnwidth]{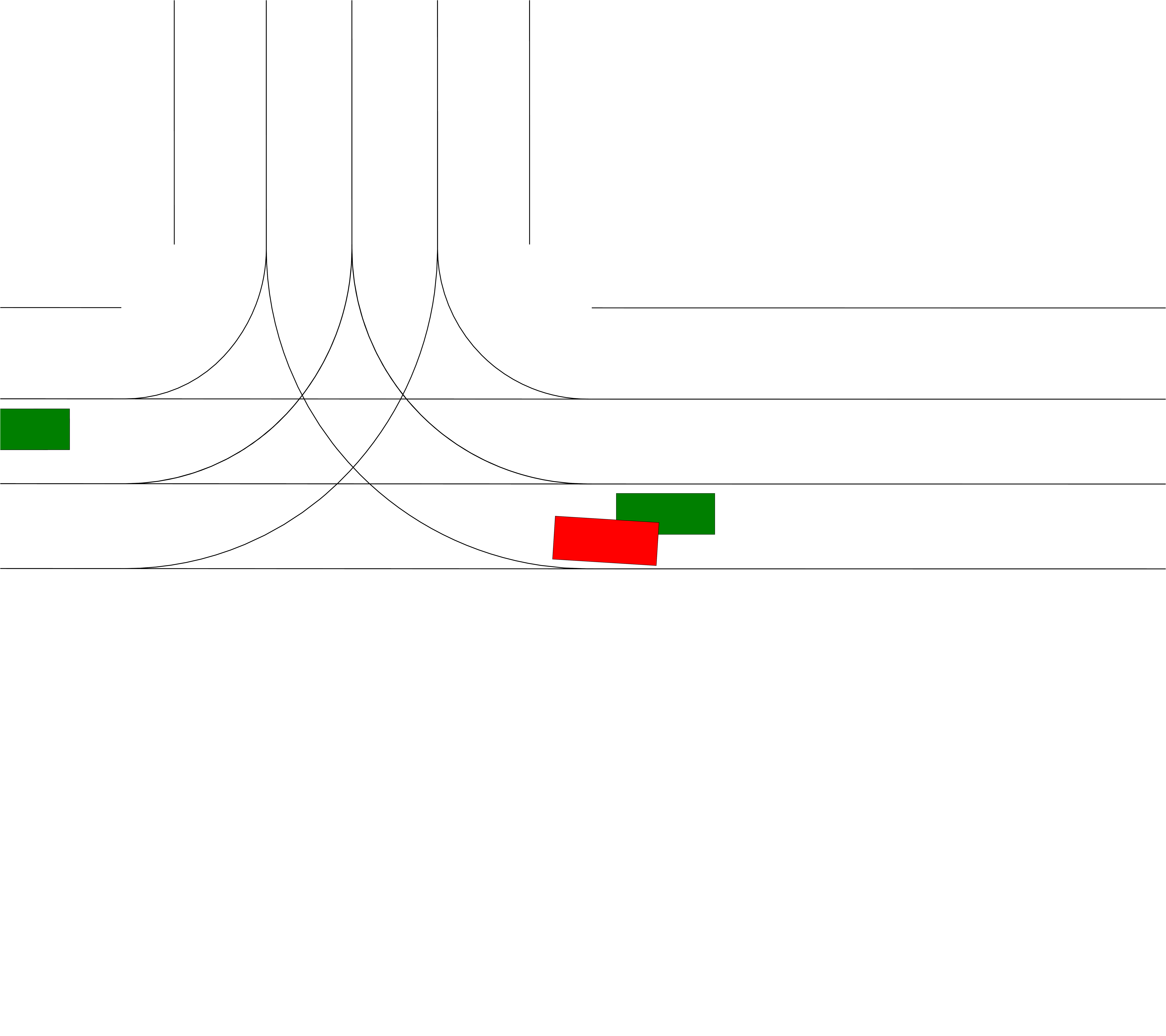} \\
        \rotatebox{90}{\hspace{0.2cm}\scriptsize ObP: right turn} &
        \includegraphics[width=0.23\columnwidth]{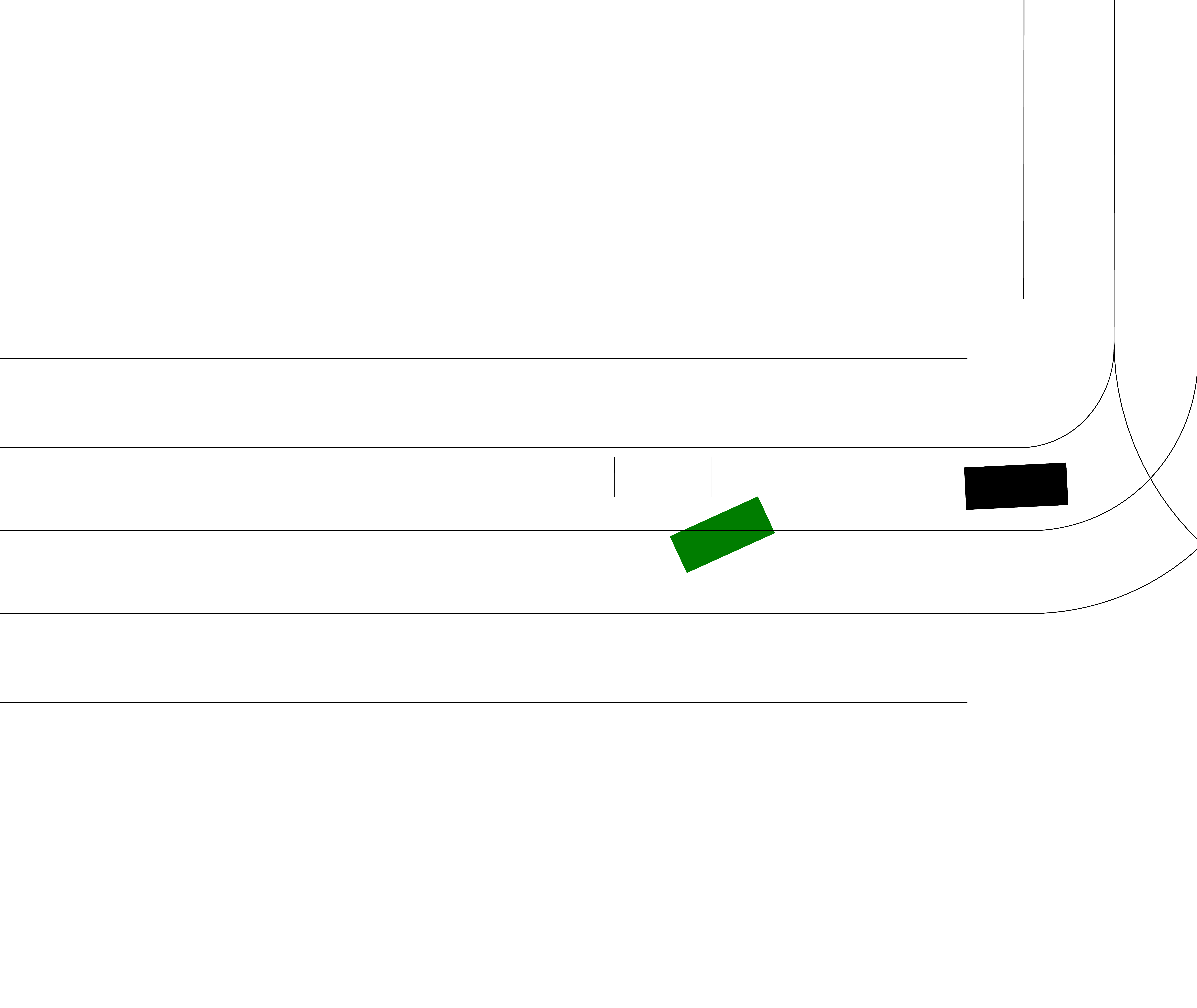} &
        \includegraphics[width=0.23\columnwidth]{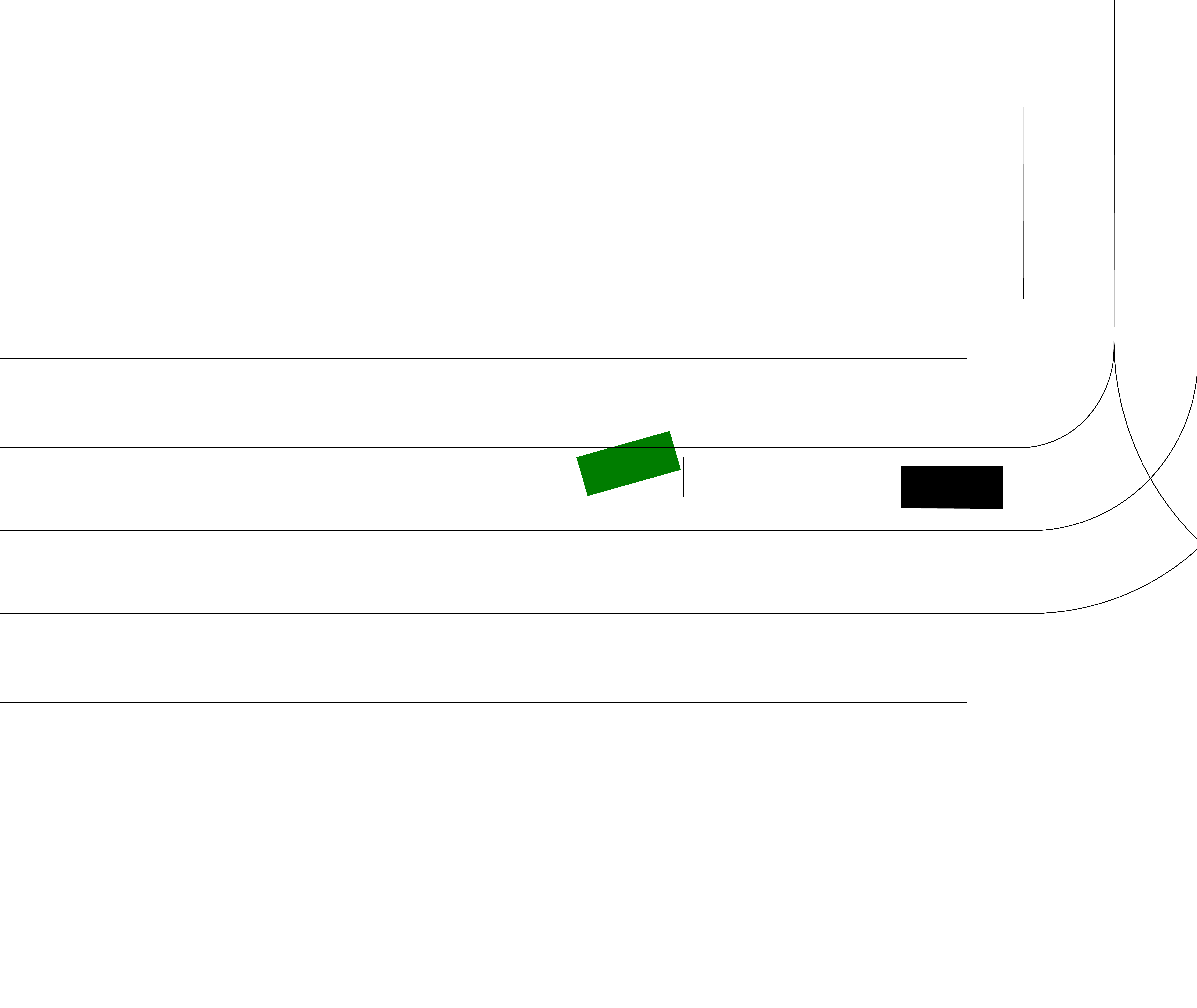} &
        \includegraphics[width=0.23\columnwidth]{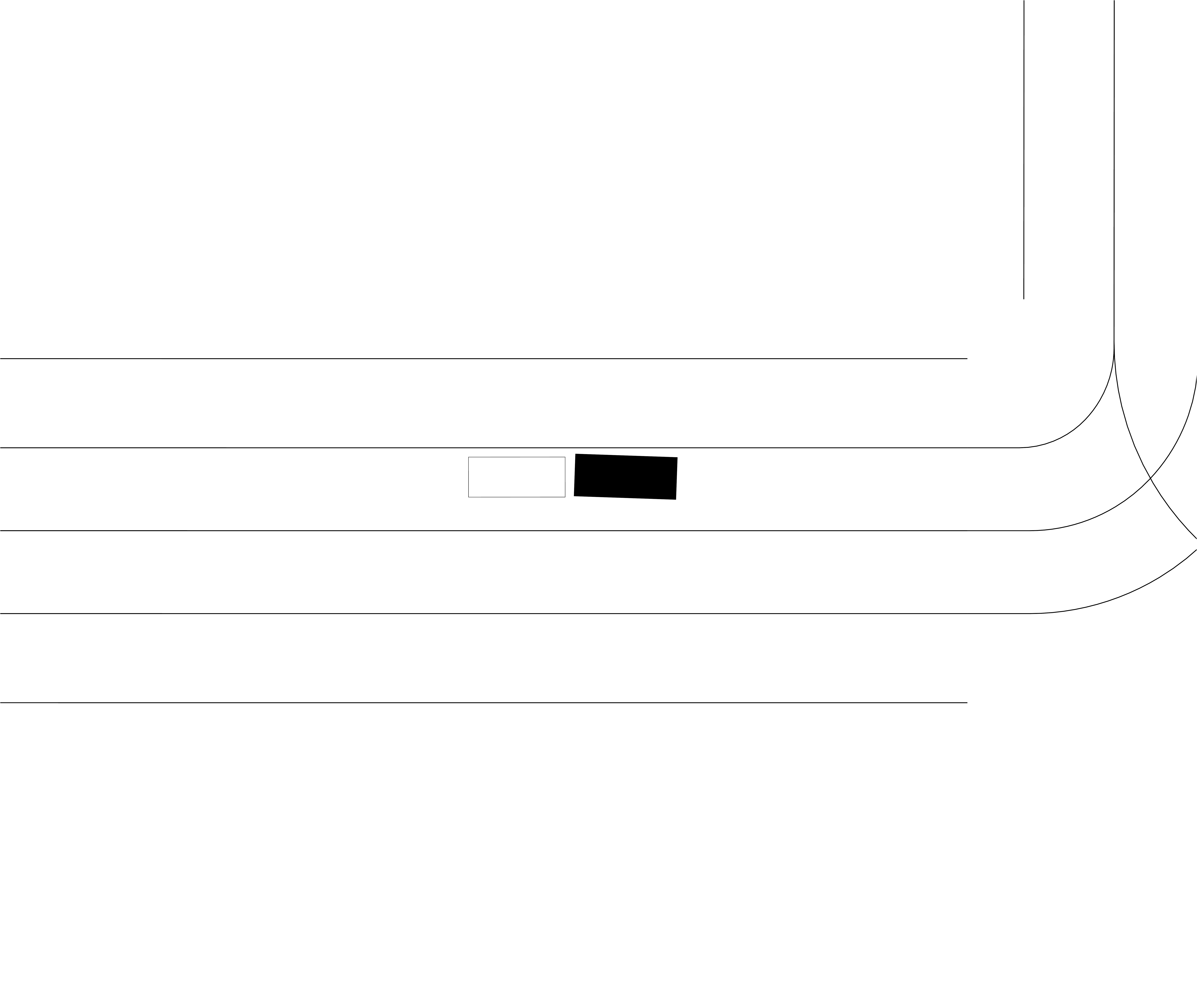} &
        \includegraphics[width=0.23\columnwidth]{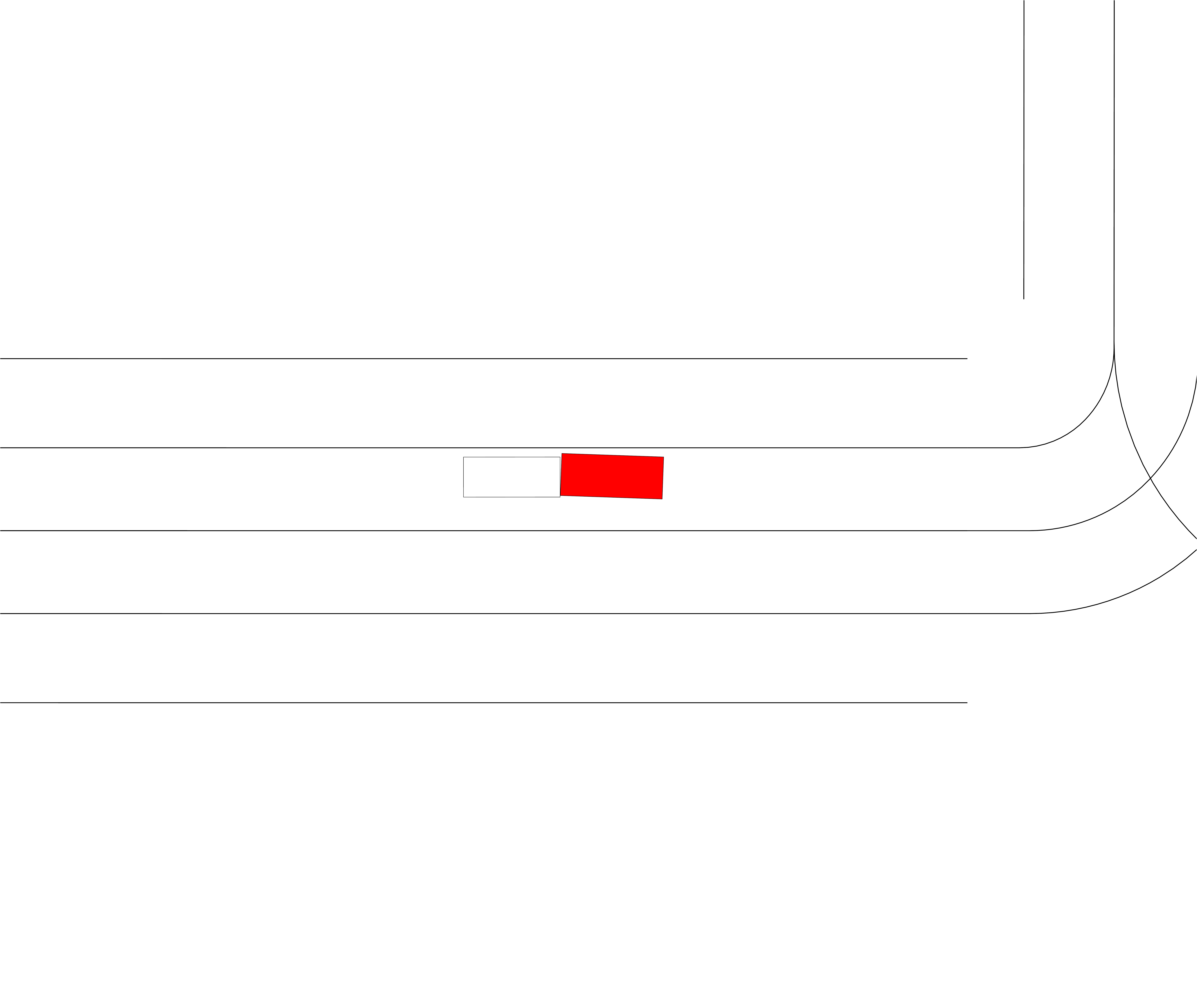} \\
        \rotatebox{90}{\hspace{0.2cm}\scriptsize ObP: overtake} &
        \includegraphics[width=0.23\columnwidth]{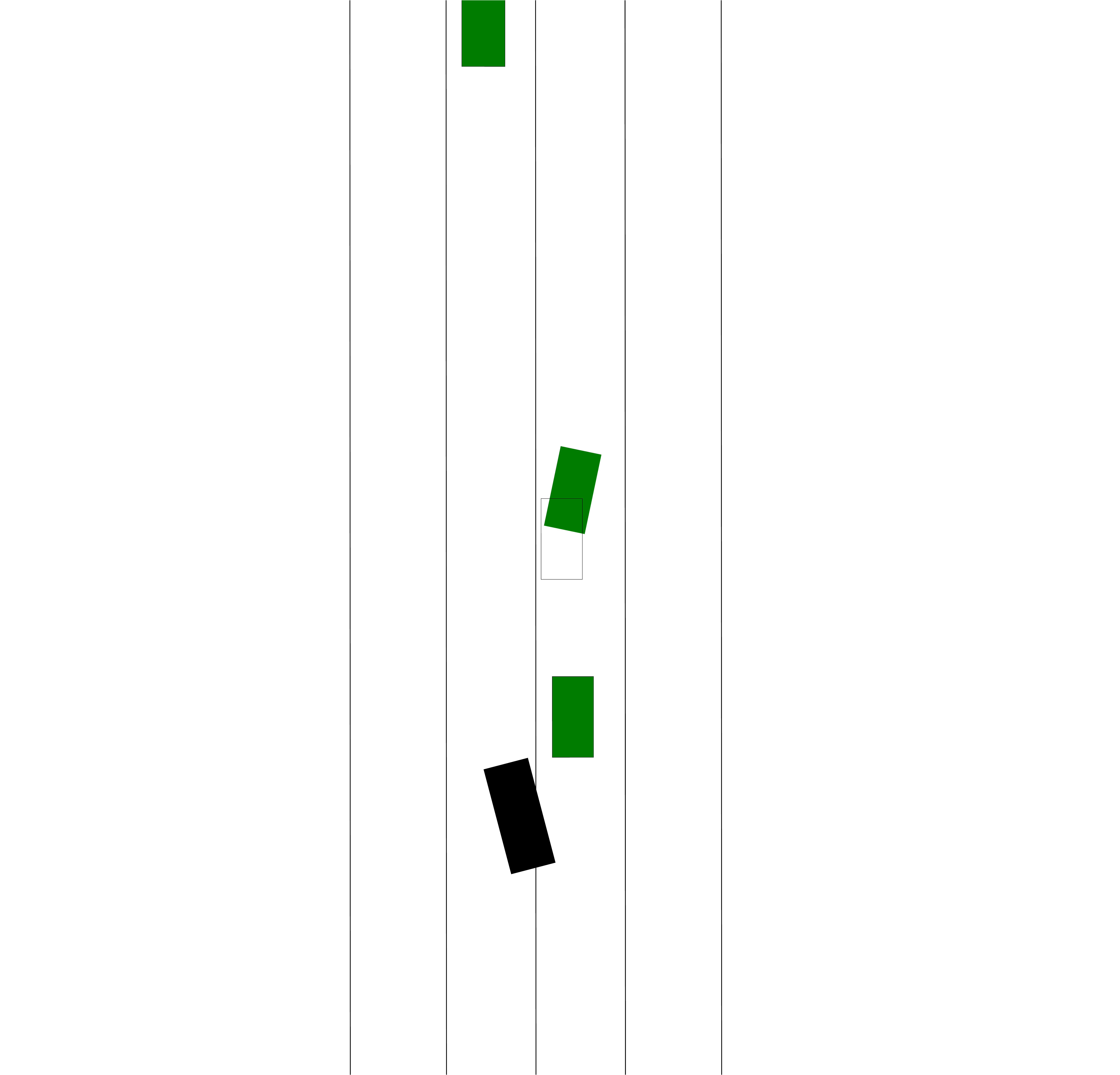} &
        \includegraphics[width=0.23\columnwidth]{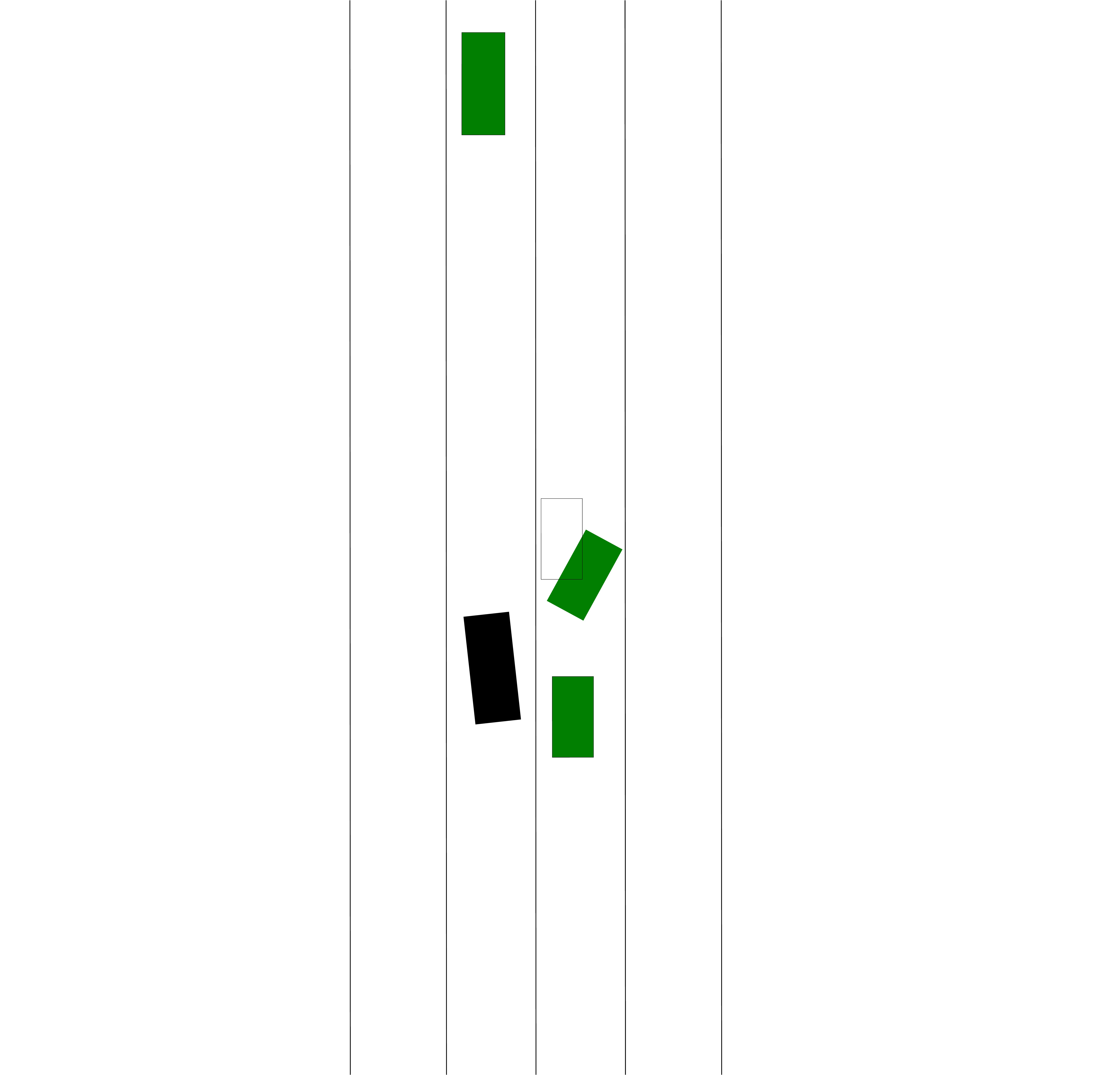} &
        \includegraphics[width=0.23\columnwidth]{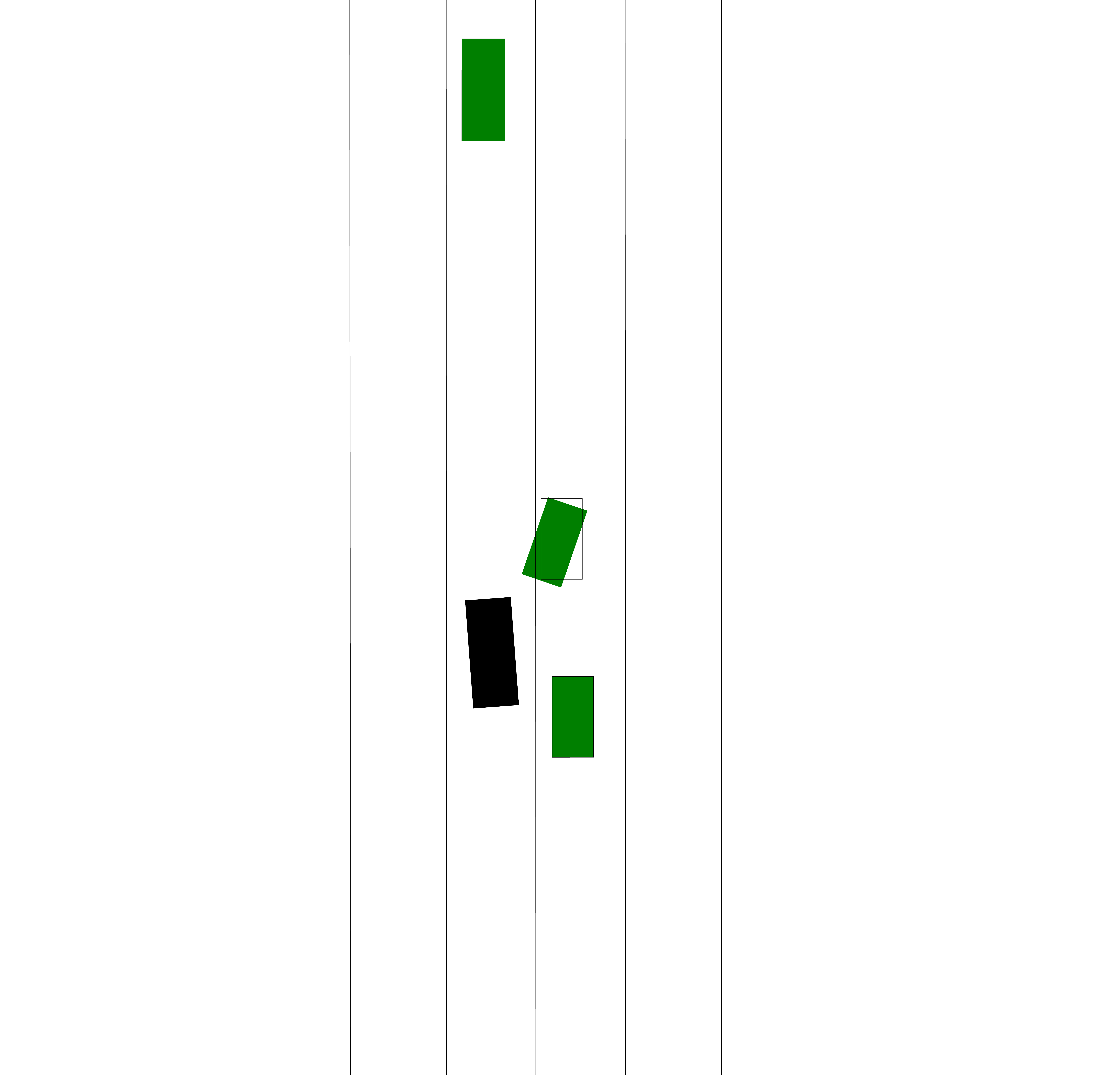} &
        \includegraphics[width=0.23\columnwidth]{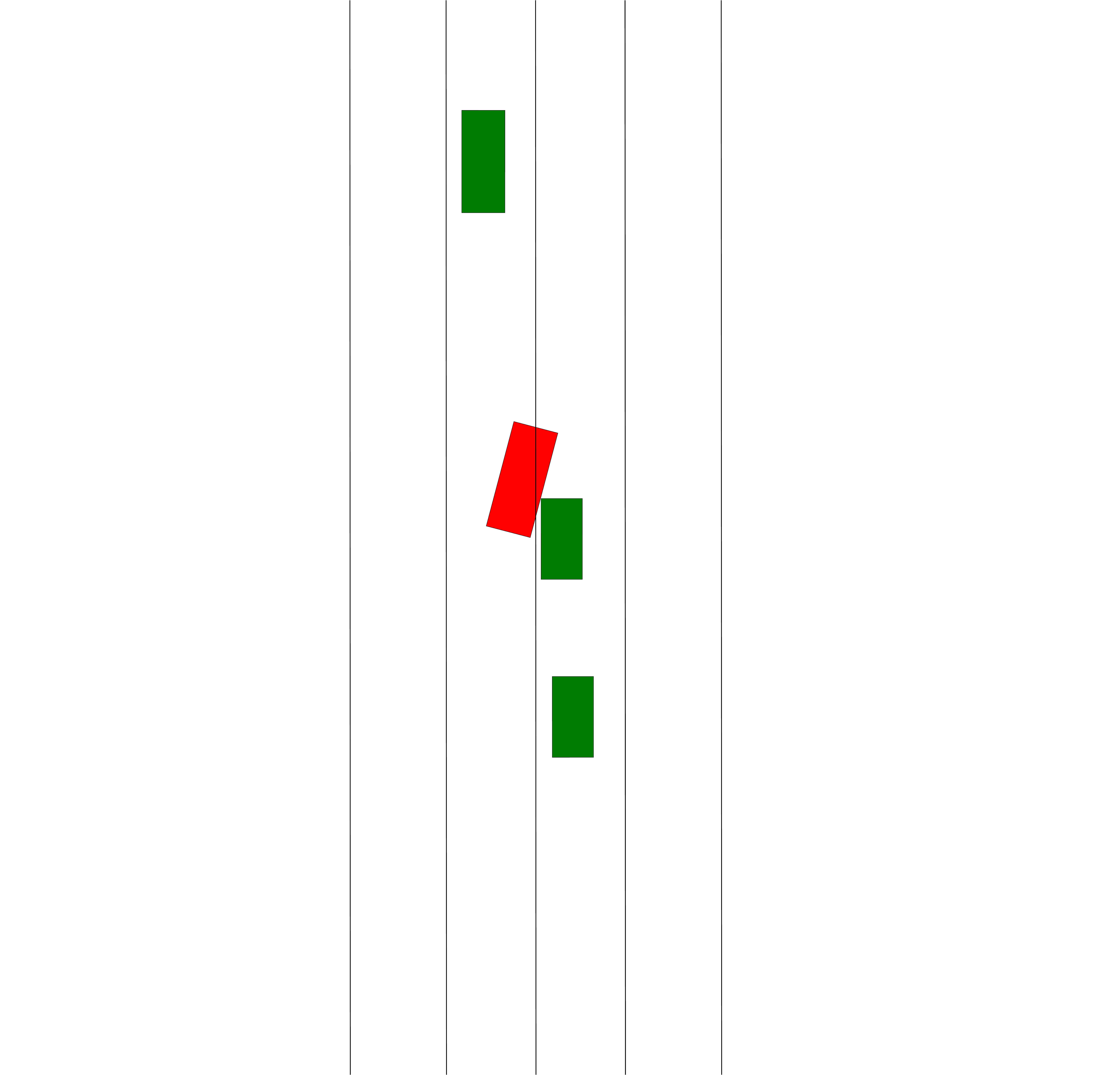} \\
        &&&& \\
        \hline \hline
        & \multicolumn{3}{l}{{\scriptsize time} $\to$} & \multicolumn{1}{c}{{\scriptsize Collision}} \vspace{.3em} \\
    \end{tabular}
    \end{minipage}
    \caption{(left) Highest values of the perception quality metrics $m$ obtained during successive rollouts of our adversarial attack search algorithm.
    Heuristic search is represented by solid lines, whilst random search is represented by dashed lines.
    We plot the $0.50$, $0.95$, and $0.99$ quantiles of the histograms of NDS/NDS-t on nuScenes val from \cref{fig:pem} for comparison.
    (right) Selected frames from adversarial attacks obtained using our algorithm, demonstrating typical errors and the resulting collision.
    }
    \label{fig:perception_metric_attack}
\end{figure*}

\subsection{Constructing adversarial perception errors}
\label{sec:experiments:constructing}

Although our system has been tuned to be robust to 
a sample set of typical
perception errors from the perception system, our algorithm is nevertheless able to produce errors of high perception quality (measured by the NDS and NDS-t metrics) that cause planning failures.
In \cref{tab:summary} we show the properties of the adversarial perception errors obtained by our algorithm.
Compared to the performance of the BEVFusion detector on the nuScenes dataset, most of the adversarial attacks score above the 99th percentile of values observed on a held-out subset of nuScenes (see \cref{fig:pem} for the full distribution).
Strikingly, most attacks maximising the NDS-t metric achieve perception scores that are above \emph{every score observed when running BEVFusion on the nuScenes validation set}.
Interestingly, we see some variability between scenarios, \eg~attacks on the \emph{lane following} scenario have lower perception score.
This scenario is arguably the simplest since it only requires the ego to adapt its velocity to other vehicles in the same lane, which could make it harder for our algorithm to find adversarial perception errors.

We analyse how adversarial perception errors are constructed by our algorithm on \cref{fig:perception_metric_attack} (left), which shows the largest perception metric $m$ achieved after a certain number of rollouts during the heuristic and random searches.
Note that each rollout on the plot is adversarial, i.e.~it leads to a planning failure.
We first observe that for both NDS and NDS-t, the heuristic search is able to find adversarial perception errors that score highly on these metrics.
We further see that in most cases the random search is able to improve these results and significantly improve the perception metrics.
The number of false negatives decreases significantly, especially when optimising the NDS-t metric which penalises contiguous false-negative detections.
Strikingly, we find adversarial perception errors with very few false negatives (2, 6, and 1) on the \emph{left turn} scenario.
It is interesting to note that the algorithm achieves this by introducing small position and orientation errors, which have a lower impact in the computation of the metric.
However the random search is not always successful in improving on the heuristic search for all scenarios, indicating that our strategy for the random search is not always effective.

On \cref{fig:perception_metric_attack} right, we show some sample frames of the adversarial perception errors obtained by maximising the NDS-t metric for the \emph{overtake}, \emph{right turn} and \emph{left turn} scenarios of the ObP planner.
For each frame, we show the ego (black/red), the ground-truth position of the other agents (white), as well as the adversarial perception (green).
We see that the perception errors are concentrated at times where the ego is close other vehicles and that vehicles that do not participate in the collision are generally perceived perfectly.
When the ego is close to the colliding vehicle, the algorithm adds both position and orientation errors and is able to achieve a collision with a very small number of false negative detections (less than 0.06\% in the \emph{left turn} and \emph{overtake} scenarios).
This is because maximising the NDS-t metric, which penalises contiguous false negatives, incentives our algorithm to trade false-negative detections for position and orientation errors during random search.
For the \emph{right turn} scenario, the adversarial attack comprises position and orientation errors at the beginning of the rollout and some false-negative detections just before collision (totalling 0.074\%).
We have also noted a pattern in which the attack targets the tracker by successively shifting the centroids of temporally clustered spatial errors.
This shows that our algorithm is able to find the right timing and combination of errors to create a planning failure while keeping perception metrics high.
We also provide videos of these adversarial attacks on all metrics and scenarios in the supplementary material.

\begin{table}
    \centering
		\begin{tabular}{p{.42in}p{.37in}p{.2in}p{.3in}p{.2in}p{.3in}p{.3in}}
		\toprule
Search type & Total FN/TP &  MPE (m) &  MAOE (rads) &  NDS &  \mbox{NDS-t} &  PEM LL \\
    \midrule
  &  \multicolumn{6}{c}{IDM: Lane Following}   \\
  \cmidrule{2-7} 
heuristic     &       172/293 &                     N/A &                                       N/A &            0.79 &   0.85 &  2.30 \\
NDS   &       148/327 &                     0.06 &                                       0.01 &             0.81 &   0.86 &  1.80 \\
NDS-t &        63/410 &                     0.55 &                                       0.05 &              0.78 &   0.88 & -1.25 \\
  \cmidrule{2-7} 
  
  &  \multicolumn{6}{c}{IDM: Overtake follow}   \\
  \cmidrule{2-7} 
heuristic     &        69/613 &                       N/A &                                       N/A &            0.94 &   0.97 & 2.14 \\
NDS   &        67/615 &                       0.00 &                                       0.00 &              0.94 &   0.97 & 2.13 \\
NDS-t &        35/674 &                       0.05 &                                       0.01 &               0.96 &   0.98 & 1.92 \\
  \cmidrule{2-7} 
  &  \multicolumn{6}{c}{ObP: overtake}   \\
  \cmidrule{2-7} 
heuristic     &        63/278 &                                 N/A &                                       N/A &             0.89 &   0.74 & 2.49 \\
NDS   &        22/706 &                                 0.10 &                                       0.01 &              0.96 &   0.97 & 2.49 \\
NDS-t &         2/339 &                                 0.21 &                                       0.06 &             0.94 &   0.97 & 1.69 \\
  \cmidrule{2-7} 
  &  \multicolumn{6}{c}{ObP: Right turn}   \\
  \cmidrule{2-7} 
heuristic     &        70/747 &                                N/A &                                       N/A &          0.95 &   0.96 & 1.66 \\
NDS   &        27/600 &                                0.02 &                                       0.01 &              0.97 &   0.97 & 1.25 \\
NDS-t &         6/807 &                                0.14 &                                       0.02 &            0.96 &   0.98 & 0.87 \\
  \cmidrule{2-7} 
  &  \multicolumn{6}{c}{ObP: Left turn}   \\
  \cmidrule{2-7} 
heuristic     &        48/330 &                                 N/A &                                       N/A &              0.93 &   0.90 & 2.67 \\
NDS   &         1/389 &                                 0.06 &                                       0.04 &             0.98 &   0.99 & 2.27 \\
NDS-t &         1/383 &                                 0.12 &                                       0.05 &        0.95 &   0.97 & 1.83 \\
         \bottomrule
    \end{tabular}
    \caption{Summary of highest obtained $m$ errors for heuristic and random searches for the NDS and NDS-t metrics. Note that we use the same heuristic search when maximising the NDS and the NDS-t metrics as initialisation for the random search. Nomenclature: MPE=Mean Position Error, MAOE=Mean Absolute Orientation Error, PEM LL=PEM log likelihood.
         }

    \label{tab:summary}
    
\end{table}

\subsection{Plausibility of the attacks}
\label{sec:experiments:properties}
\begin{figure}
    \centering
    \includegraphics[scale=1]{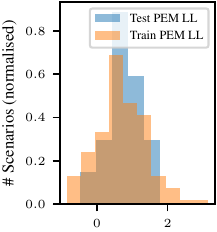}
    \hfill
    \includegraphics[scale=1]{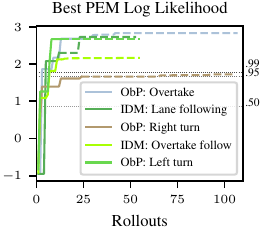}
    \caption{
    Adversarial attacks that maximise the PEM log likelihood (LL).
    (left) Distribution of the PEM LL on the nuScenes val dataset for the BEVFusion detector. The nuScenes val dataset was split in a 0.9:0.1 ratio to create train and test datasets for the PEM which are themselves independent to the nuScenes train dataset used to train BEVFusion.
    (right) Highest PEM LL achieved obtained by the heuristic and random searches.
    The solid line represents heuristic search. The dashed line represents random search.
    }
    \label{fig:pem results}
\end{figure}

In \cref{tab:summary} we show the PEM log likelihoods of the attacks (PEM LL) and see that they rank in the 95--99th percentiles compared to PEM LL values obtained on a held-out subset of nuScenes, see \cref{fig:pem results} left.
This means that the errors obtained by our algorithm are 
relatively high-likelihood
according to the PEM.
It is interesting to note that maximising the NDS and NDS-t metrics further using the random search does reduce the PEM LL consistently compared using the heuristic search alone, indicating that adversarial errors become less 
likely
as they become more specific.
It is also possible to use random search to maximise the PEM LL directly, see \cref{fig:pem results} left and \cref{tab:summary_pem} in the appendix for detailed results.
In this case, we obtain attacks that rank in the 99th percentile of the PEM LL for all scenarios except \emph{right turn}. Note that this comes with only a modest decrease in NDS and NDS-t which means that these attacks both are likely according to the PEM and score highly on the NDS and NDS-t metrics.
We would of course be unable to exclude the possibility of such errors occurring in the real world on the basis of these figures alone.

\begin{figure}
    \centering
    \includegraphics[width=0.49\columnwidth]{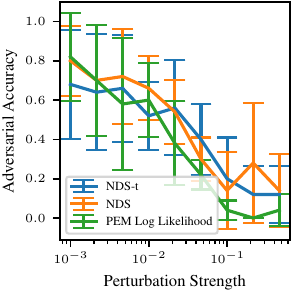}
    \hfill
    \includegraphics[width=0.49\columnwidth]{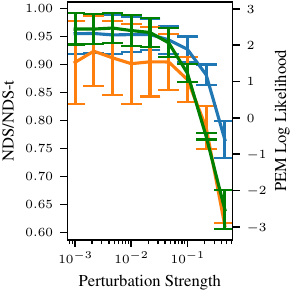}
    \caption{Change in the fraction of adversarial rollouts (left) and perception metrics (right) when perturbations of increasing strength are applied to adversarial perception errors obtained in previous section. Average is taken over 10 random perturbations and all scenarios. The perception metrics decrease more slowly than the adversarial accuracy, indicating that perception error sequences with similar perception metric values can cause very different behaviour with respect to planning rules.
    }
    \label{fig:robustness}
\end{figure}

However, we are also interested in the effective support of a given attack: that is, over how large a region of error space a given example remains adversarial.
To investigate this, we probe the immediate neighbourhood of the adversarial attacks by applying random perturbations of increasing strengths to the adversarial perception errors obtained in the previous section and check if the resulting perception inputs still cause a planning failure.
The random perturbations consist of randomly flipping detections to be false negatives or true positives with a specific probability and adding random spatial noise.
In \cref{fig:robustness} we show how the fraction of adversarial scenarios (adversarial accuracy) and the average perception quality behave for increasing perturbation strength, which we take to be both the flip probability and the standard deviation of the spatial noise.
We observe that the percentage of rule-breaking rollouts decreases quickly as the strength of the perturbation increases, while the perception quality stays high for longer.
This indicates that there are non-adversarial perception inputs of high perception quality around adversarial perception errors, i.e.~that adversarial attacks are relatively isolated in this error space.

\section{Discussion}

\paragraph{\emph{Are adversarial perception errors likely to occur in the real world?}}

The adversarial perception errors are judged to be both high-likelihood by the PEM and high-quality under NDS/NDS-t.
From this perspective, they appear to be inliers, \emph{other} than in their particularly detrimental effect on the planner.
This is unsettling on its face, as these appear to be ``likely failures''.
Yet, they were not sampled when tuning the stack's hyperparameters (see \cref{sec:experiments:setup}).
This turns our attention back to the extent of support of such events: their probability of occurrence is not given directly by the likelihood model itself, but by its integral over the subset of the domain corresponding to failure-triggering outputs.
The support of interest here is the intersection of the respective subsets of perception output space over which the planner constraint is violated and the perception quality score exceeds a threshold.
In practice, neither this domain nor the likelihood integral over it can be computed.

This is in fact one of the concerns we wish to raise.
We have demonstrated that a simple optimisation method can locate positive-measure regions of perception output space whose likelihoods are not only non-zero, but \emph{high}.
Therefore, while we cannot easily establish that the total probability mass of such events exceeds a given safety threshold, nor can we establish that it does \emph{not}.
As in \cref{sec:experiments:properties}, we can take steps towards estimating the support of a given example once it is located, but we are well short of a reasonable estimate of the aggregate probability mass of concern.

In that vein, we further note that potential criticism of disparity between the PEM and the true error distribution would carry over to sampling-based attempts to establish probabilistic performance guarantees.
While likelihood is, as above, relevant to the question of the probability that an event will be observed in the wild, our active approach decouples the identification of adversarial perception errors from the estimation of their probability mass.
We consider this to offer benefits over sampling-based approaches when the error model cannot be fully trusted (as it never can be).
Likewise, if the produced examples are subjectively judged to be of low perception quality whilst scoring high on NDS/NDS-t, then this reveals a critical limitation of the perception score metrics themselves.

\paragraph{\emph{Are adversarial perception errors useful?}}

Adversarial perception errors can be surprisingly interpretable.
That is, they can reveal underlying algorithmic weaknesses in the planner.
\cref{sec:experiments:constructing} points out examples of this, which can be viewed in the supplementary videos.
These and other examples may come as a surprise to the practitioner, and may provide useful information for refining the planner (manually if necessary).
In general, we view this tool as fitting into the ``fuzzing'' paradigm of software testing.
We advocate the inclusion of worst-case analysis in any practitioner's overall approach.

\section{Conclusion}

In this work, we proposed a novel framework for defining and identifying erroneous perception system outputs which cause failures in modular autonomous vehicles with widely used components.
Surprisingly, these failures occur despite the identified perception errors appearing benign when analysed with common perception metrics.
Key to our success is a modified version of the Boundary Attack algorithm, which uses a combination of heuristic and random search to identify these failures for black-box driving systems and simulators that do not provide gradients. 
We provide experimental results to demonstrate that our approach works well in practice on a number of driving scenarios that are relevant to the industrial deployment of autonomous vehicle systems.
We hope that this work opens possibilities to further explore and evaluate the downstream behaviour of planner components in light of mistakes (perception errors) made by upstream components in autonomous vehicles.

\section*{Acknowledgements}
The authors wish to express their gratitude to all present and former Five employees who have contributed to internal planning and perception software which enabled this research.
In particular we wish to thank Mihai Dobre for offering comments on this manuscript, and Ludovico Carozza for providing technical support with planning software and the CARLA simulator.

{
    \small
    \bibliographystyle{ieeenat_fullname}
    \bibliography{refs}
}

\clearpage
\maketitlesupplementary

\setcounter{section}{0}
\setcounter{table}{0}
\setcounter{figure}{0}

\renewcommand{\thesection}{\Alph{section}}
\renewcommand{\thetable}{S\arabic{table}}
\renewcommand{\thefigure}{S\arabic{figure}}

\section{Definition of metrics}
\label{sec:metrics}
The nuScenes detection score (NDS) is defined as
\begin{equation}
    \textrm{NDS} = \frac{1}{10} [5~\textrm{mAP} + \sum_{\textrm{mTP} \in \mathbb{TP}} ( 1 - \min(1, \; \textrm{mTP}) ) ] ,
\end{equation}
where $\textrm{mAP}$ is the mean average precision, and the metrics defined on true positive boxes are defined as $\textrm{mTP} = \frac{1}{|\mathbb{C}|} \sum_{c \in \mathbb{C} } \textrm{TP}_{c}$, where the average is taken over all classes $c \in \mathbb{C}$ \cite{caesar2020nuScenes}.
The true positive metrics are: Average Translation Error (ATE), Average Scale Error (ASE), Average Orientation Error (AOE), Average Velocity Error (AVE), and Average Attribute Error (AAE).
In our experimental setup AAE is not used so this error metric is set to the minimum value (0).

NDS-t is defined as
\begin{equation}
    \textrm{NDS-t} = \frac{\textrm{NDS} + (1 - \textrm{Longest Drop Fraction})}{2},
\end{equation}
where $\textrm{Longest Drop Fraction}$ is defined as the longest fraction of any track that is a continuous false negative.

\section{Perception Error Model}
\label{sec:pem}
\subsection{Background}

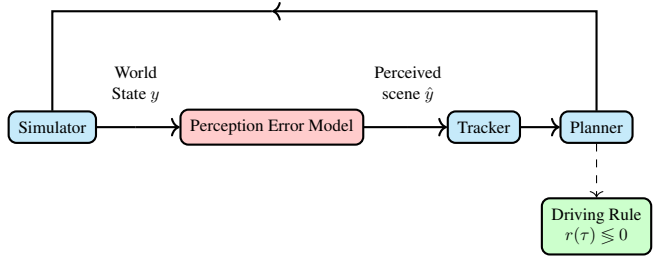
\begin{figure}
    \centering
    \begin{tikzpicture}[node distance=0.8cm,scale=0.65]
\tikzset{every node/.style={transform shape, align=center, draw, rounded corners=3, inner sep=2mm, thick}}

\node[fill=cyan!20] (simulator) {Simulator};
\coordinate[right=of simulator] (c1);
\node[right=0.9cm of c1,  fill=pink!80] (attack) {Perception Error Model};
\coordinate[right=0.9cm of attack] (c2);
\node[right=of c2, fill=cyan!20] (tracker) {Tracker};
\node[right=of tracker, fill=cyan!20] (planner) {Planner};
\node[fill=green!20] at (search-|planner) (eval) {Driving Rule\\$r(\tau) \lessgtr 0$};
\coordinate[above=2cm of attack] (top);

\draw[thick] (simulator) -- (c1);
\draw[->, thick] (c1) -- (attack);
\draw[thick] (attack) -- (c2);
\draw[->, thick] (c2) -- (tracker);
\draw[->, thick] (tracker) -- (planner);
\draw[->, thick] (planner.north) |- (top);
\draw[thick] (simulator.north) |- (top);
\draw[->, dashed] (planner) -- (eval);

\node[draw=none, above=0.4cm of c1, fill=white, inner sep=1mm] {World\\State $y$};
\node[draw=none, above=0.4cm of c2, fill=white, inner sep=1mm] {Perceived\\scene $\hat y $};

\end{tikzpicture}
    \caption{System configuration when testing with Perception Error Model.}
    \label{fig:pem_system}
\end{figure}

Perception error models (PEM) can be used to approximate $f$ using a probability distribution conditioned on an \emph{augmented} state $\tilde{s}$, which is cheaper to produce in simulation than the actual state $s$, because expensive-to-compute sensor data is not included in $\tilde{s}$ \cite{innes2023testing, sadeghi2021step}.
The approximation is probabilistic because the augmented state does not include all the information required to predict the perceived state $\hat{s}$ exactly.
To simplify our notation we denote the PEM as a distribution of perceived states, $p(\hat{s}\mid s)$, conditioned on $s$.
The simulation pipeline when using a PEM is shown in \cref{fig:pem_system}.
When simulating with the PEM, the probability of transitioning to state $s'$ from $s$ is
\begin{align}
p(s', a \mid s) 
&=  p(s' \mid s, a) p(a\mid s) \nonumber \\
&= p(s' \mid s, a) \int \delta(a - \pi(\hat{s})) p(\hat{s} \mid s) d \hat{s}.
\end{align}
Then starting from $s_0$ we define a $T$-step simulation rollout as $\tau = \left[ s_0, a_1, s_1, a_2 \dots s_{T - 1} \right]$, where the rollout probability is:
\begin{multline}
      p(\tau) = \prod_{t=1}^{T-1} p(s_t \mid s_{t-1}, a_{t}) \\ \times \int \delta(a_t - \pi(\hat{s}_{t-1})) p(\hat{s}_{t-1} \mid s_{t-1}) d \hat{s}_{t-1}.
\end{multline}

\subsection{Methodology}

\label{sec:pem_implementation}

The PEM is parameterised by a neural network which factorises over each agent in each scene.
The architecture of the network consists of 5 residual blocks with tanh activations, where every layer is fully connected as in \cite{sadeghi2021step}.

To create training data for the PEM, a tuple $\{s, \hat{s} \}$ of input-output is created for every frame by running the perception system $f$ on a labelled sensor dataset, which we process with an association algorithm to obtain an input-output tuple for each agent in the scene, and therefore the training data for the surrogate detector would be $\dataset = \{x_i , \hat{x}_i\}_{i=1}^k$ \cite{sadeghi2021step, innes2023testing}.
The inputs to the neural network are the position, extent, yaw, and percentage occlusion of the agent concatenated with a one hot encoding of the object class of the object.
We model occlusion levels by running a lightweight ``low-fidelity rendering'' in order to obtain percentage occlusion for each object, which we describe in greater detail in \cref{sec:low_fid}. 

The PEM is trained by optimising the parameters of a probabilistic neural network to minimise the loss
\begin{equation}
    \mathcal{L}_\text{total} = \sum_i \log p({\hat{x}_i}^\text{det}|x_{i}) + \indicator_{\{\hat{x}_i^\text{det} = 1\}} \log p(\hat{x}_i^\text{pos}| x_{i}),
    \label{eqn:loss}
\end{equation}
where $p(\cdot|x_i)$ represents the likelihood, $\hat{x}_i^\text{det}$ represents the Boolean output which is true if the object was detected, and $\hat{x}_i^\text{pos}$ represents a real-valued output describing the centre position of the detected object, respectively.
The term $\log p(\hat{x}_i^\text{det}|x_i)$ in \cref{eqn:loss} is equivalent to the binary cross-entropy when using a Bernoulli distribution to predict false negatives. 
In this paper we make a slight departure from previous work; the position error is not predicted by independent normal distributions, but instead by a Multivariate Student T distribution which enables a more accurate characterisation of errors.
The log likelihood of the multivariate student T distribution is used for $\log p(\hat{x}_i^\text{pos}| x_{i})$, which is parameterised by a location vector, scale matrix and scalar degrees of freedom which are the outputs of the fully connected neural network --- an implementation of the distribution is available in Pyro \cite{bingham2019pyro}.
Of course, similar loss functions can be defined for many different distributions and model architectures.

When training, we set the dropout probability to $0.2$ to prevent overfitting.
The batch size was 10000. 
The learning rate for the adam optimiser was $10^{-3}$.
We train for 1000 epochs.

We can evaluate the PEM by comparing properties of samples from $p(\hat{s}\mid s)$, such as mean average precision, with the same properties of the original perception system outputs.

\subsection{Experimental Results}
\label{sec:pem_nds}
\begin{figure}
    \centering
    \begin{subfigure}{0.48\columnwidth}    \includegraphics{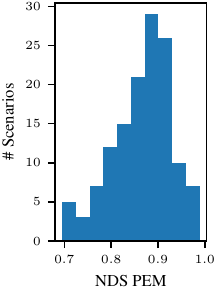}
    \caption{NDS Train}
    \end{subfigure}
        \begin{subfigure}{0.48\columnwidth}     \includegraphics{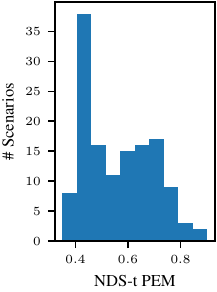}
    \caption{NDS-t Train}
    \end{subfigure}
        \begin{subfigure}{0.48\columnwidth}     \includegraphics{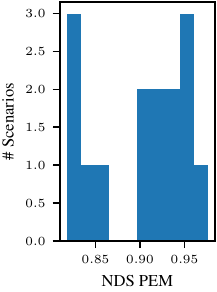}
    \caption{NDS Test}
    \end{subfigure}
        \begin{subfigure}{0.48\columnwidth}     \includegraphics{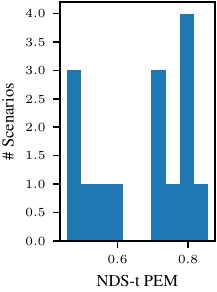}
    \caption{NDS-t Test}
    \end{subfigure}
    \caption{Performance of PEM on perception metrics for train and test split for PEM training of nuScenes val dataset.}
    \label{fig:pem_nds}
\end{figure}

\cref{fig:pem_nds} shows an analysis of samples from the PEM with the NDS and NDS-t metrics on the nuscenes val dataset, using the train/test split which was used to train the PEM.
Overall the shape of these histograms is similar to those in \cref{fig:pem}, indicating a good agreement between the output of the PEM and the training/test data according to NDS and NDS-t.

In \cref{tab:summary_pem} we show the properties of the attacks obtained by maximising the PEM LL directly with random search.
We observe that in many cases the number of false negatives is higher than for the attacks on NDS and NDS-t in \cref{tab:summary}, perhaps indicating that the PEM LL does not penalise some false negatives.
However, compared to \cref{tab:summary} we notice that there are smaller position errors, indicating that these are perhaps penalised more severely.
Although the NDS and NDS-t is slightly reduced compared to \cref{tab:summary} and PEM LL is slightly increased, most of these values are still high compared to the histograms on the nuScenes val set shown in \cref{fig:pem} and \cref{fig:pem results}.

\begin{table*}
    \centering
    
    \caption{Summary of highest obtained PEM log likelihood errors with random search. 
         }
		\begin{tabular}{p{1in}p{.3in}p{.3in}p{.5in}p{1in}p{.2in}p{.4in}p{.7in}}
		\toprule
Scenario &   Total FN &  Total TP &   Mean Position Error (m) &  Mean Absolute Orientation Error (radians) &   NDS &  NDS-t &  PEM  Log Likelihood \\
    \midrule
IDM: Lane Following   &      145 &       358 &                                  0.00 &                                             0.00 & 0.84 &   0.88 &  2.73 \\

IDM: Overtake follow  &        60 &       622 &                                      0.00 &                                             0.00 & 0.95 &   0.97 & 2.17 \\

ObP: overtake   &          40 &       211 &                               0.01 &                                                 0.25 & 0.91 &   0.81 & 2.51 \\

ObP: Right turn   &       41 &       774 &                                  0.01 &                                                 0.00 & 0.97 &   0.98 & 1.72 \\

ObP: Left turn   &        47 &       331 &                                  0.00 &                                                0.00 & 0.93 &   0.90 & 2.68 \\
         \bottomrule
    \end{tabular}
    \label{tab:summary_pem}
\end{table*}

\subsection{Low Fidelity Occlusion Rendering Approach}
\label{sec:low_fid}

\begin{figure}
    \centering
    
\resizebox{1\columnwidth}{!}{%
    \begin{tikzpicture}
\draw  (27.75,26.5) rectangle  node {\LARGE Ego} (29.25,23.5);
\draw [, dashed] (28.5,25) circle (4.25);
\draw [, dashed] (28.5,25) circle (8);
\draw  (35.5,32) rectangle (37,29);
\draw  (34,29.5) rectangle (32.5,26.5);
\draw  (19,26.5) rectangle (20.5,23.5);
\draw[-latex, thick] (29.75,25.25) --  (32,26.5);
\node [font=\LARGE] at (30.5,26) {r};
\draw [stealth-stealth, thick] (31.25,27.75) .. controls (32,26.75) and (32,26.75) .. (32.5,25.75);
\node [font=\LARGE] at (31.8,25.2) {[$\underline{\theta}$, $\bar{\theta}$]};
\draw[blue, ultra thick] (32.68,25.738) arc (10:55:4.25cm);
\draw[blue, ultra thick] (20.621,26.389) arc (170:190:8cm);
\draw[blue, ultra thick] (36.0175,27.736) arc (20:48:8cm);
    
\end{tikzpicture}
    }
    \caption{Diagram showing how a birds eye view representation of vehicles in the ego frame can be converted to an arc segment representation. The vehicles displayed as boxes are converted into the arc segments, shown in blue.}
    \label{fig:occlusion}
\end{figure}
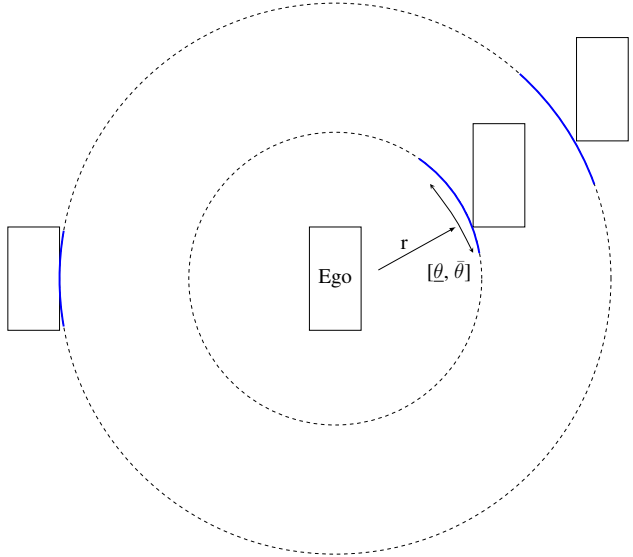

In order to estimate the percentage occlusion for each agent in the scene we convert the birds eye view representation of the scene to a radial birds eye view representation in ego coordinates by converting each agent into an arc segment in radial space which bounds the minimum distance of the agent from ego and bounds the angular coordinate from ego.
Then we sort the arcs by distance from ego and check for overlap between each arc segment and all closer arc segments.
The percentage occlusion is calculated as the largest percentage of the arc segment which is intersected by any closer arc segment.
Since the calculated value is not guaranteed to be equal to the percentage occlusion which would be calculated by any individual sensor using a high fidelity 3D calculation, we term our calculated quantity \emph{low fidelity pseudo occlusion}.
A diagram of this procedure is shown in \cref{fig:occlusion}.

\section{Extended Related Work}
\label{sec:related_full}
\subsection{End-to-end evaluation in simulation} 
Several works have considered the end-to-end evaluation of safety-critical machine learning pipelines using simulated data.
Such approaches often try to scale up the number of evaluations by using a lower fidelity simulator whilst maintaining good enough accuracy in order to capture realistic failure cases, see e.g.~\cite{pouyanfar2019roads, balakrishnan2020closing, el2019LiDAR}.
More complex algorithms are also possible, for example in \citet{dennis2020emergent}, where the authors describe an approach to create progressively more difficult curricula of scenarios to optimally train an agent.
Likewise, \citet{urtasun} attempt to alter actual LiDAR data in order to find adversarial scenarios for autonomous driving systems.
\citet{kadian2019we} attempt to validate a simulator by demonstrating that the behaviour of an end-to-end point navigation network in the simulated environment mimics its real-world behaviour. 
End-to-end testing is also possible without a simulator, by considering the impact of detector outputs on a planner at a single point in time \cite{philion2020learning}.
Similarly, \citet{9981309} describes an approach to identify agents for which the outputs of an object detector are both incorrect and of high consequence, by filtering data based on a reachability analysis and typical detection performance evaluations.
\citet{corso2022risk} describe how the loss used to train a perception system can be augmented with a loss representing subsequent mistakes made by a planner acting on the output of the perception system, and hence the perception system can be specifically tuned to reduce errors which result in adverse downstream behaviour.

\subsection{Perception error models (PEMs)} 
Perception Error Models (PEMs) have been used in simulation to mimic the outputs of perception systems to enable the realistic assessment of downstream tasks.
For example, \citet{piazzoni2020modeling} propose a PEM which factors in weather conditions when determining the error distribution associated with the pose and class of agents, which is used to validate an autonomous vehicle system in a simulated urban driving scenario.
PEMs can also be used to predict false negative detections \cite{piazzonimodeling}.
Many PEMs are sequential probabilistic models \cite{berkhahntraffic, hirsenkorn2016virtual, zec2018statistical, mitra2018towards}, and some utilize modern machine learning methods \cite{krajewskineural, arnelid2019recurrent, suhre2018simulating}.
In \citet{sadeghi2021step}, the behaviour of a neural-network-based PEM in the CARLA simulator is studied in a large-scale urban driving simulation and the behaviour compared to an object detector and simulation with ground truth perception.
In \citet{innes2023testing} a PEM is deployed in an emergency braking scenario in the CARLA simulator, and an efficient importance sampling strategy is used to reduce the number of samples required to estimate the probability of collision.
\citet{philipp2021simulation} consider a low-dimensional perception error model and verify the amount of perceptual error which can be applied without a collision occurring in a simple scenario.
\citet{reeb2022validation} describe an alternative to PEMs, where instead the input and output distributions of each component in a modular system under simulation are bounded relative to their real-world distributions, and hence the behaviour of the system as a whole is simulated truthfully.

\subsection{Failure mode identification and assessment}
Many simulation techniques used in autonomous vehicle testing were used extensively in other contexts prior to the advent of autonomous vehicles.
The seminal works of \citet{hasofer1974exact} and \citet{rackwitz1978structural} introduced the concept of the design point and reliability index into reliability analysis, whereby a first-order approximation of the performance of a system is used to determine the most likely failure mode of the system, which in turn determines the failure probability of the system.
More general results are described in \citet{Breitung1994} and \citet{hohenbichler1984asymptotic}.
\citet{moller2004fuzzy} and \citet{moller2000fuzzy} describe attempts to evaluate the reliability of a system when uncertain system variables are modelled by fuzzy sets, which is similar to the use of a perception metric to specify a level of performance for the perception system used in our work.

Some works aim to directly calculate the associated probability of the failure modes; for example, in \citet{uesato2018rigorous} an efficient importance sampling approach is applied to calculate the failure probability of reinforcement learning agents, \citet{inatsu2021active} use a Gaussian process to optimise system designs to minimise the probability of failure, and \citet{sadeghi2022bayesian} use a Gaussian process surrogate model to efficiently estimate the failure boundary for an autonomous driving system when rule functions are partially defined.

Failure mode inputs of RGB-image-based deep learning systems which appear benign to humans but cause unexpected behaviour of the system are often described as \emph{adversarial}  \cite{szegedy2013intriguing}.
Such attacks may be performed in the real world \cite{ wu2020making, lee2019physical, li2019adversarial, van2019fooling, sharif2016accessorize, chen2019shapeshifter, kurakin2018adversarial,brown2017adversarial,eykholt2018robust,song2018physical}, including on autonomous vehicles \cite{morgulis2019fooling,zhang2018camou,sitawarin2018darts}.
In the `decision based' setting, where only the predicted class of the classifier is known, an algorithm was proposed by \citet{brendel2018decisionbased} to identify these failure modes. 
Furthermore, more efficient iterations of this algorithm have been developed \cite{Dong_2019_CVPR, 9152788, cheng2019query, sitawarin2022preprocessors}.
The concept of adversarial attacks has also been applied to autonomous driving by \citet{Bahari_2022_CVPR}, where synthetic road layouts are manipulated to cause driving failures.
Similarly, \citet{Roelofs2022CausalAgentsAR} describe how removing perceived agents from a scene can cause large changes in the output of trajectory prediction systems.

The identification of useful and representative scenarios which can be used to effectively test autonomous vehicles, whilst not necessarily appearing benign to humans, has emerged as a separate task from the overall estimation of failure probability for the system.
\citet{corso} provide a state-of-the-art review of black-box techniques to find safety-critical scenarios.
Similarly, \citet{zhang2021finding} provide a state-of-the-art review of methods used to identify safety-critical scenarios.
Surrogate models are used in simulation to efficiently search for failures in autonomous driving systems; for example, \citet{sinha2020neural} use a combination of efficient sampling and a surrogate model to identify failure modes and find their rate of occurrence, 
\citet{beglerovic2017testing} identify failure cases for an autonomous vehicle using Bayesian optimisation, and \citet{vemprala2021adversarial} demonstrate how adversarial scenarios can be identified for optimization based planners using Bayesian optimisation.
In \citet{corso2019adaptive}, \citet{koren2018adaptive} and \citet{koren2019efficient} various methods are proposed using a reinforcement learning solver to find the most likely failure mode of a system which is tested in an environment modelled as a Markov decision process.
\citet{hanselmann2022king} use a differentiable physics model to obtain approximate gradients of a safety rule with respect to the position of other agents in a simulation and hence efficiently obtain adversarial scenarios.

\section{Experimental Hyperparameters}

In the heuristic search we limit the bisection algorithm to at most 3 iterations every time it is called to identify $t_\text{start}$ and $t_\text{end}$.
In the random search we set $N_\text{steps}=40$ and $N_\text{proposal-steps}=100$.
The proposal distribution flips the false negative property of a track segment with uniformly distributed start and end time.
The flipped track segment will then be assigned a position error with uniformly distributed direction and uniformly distributed magnitude between 0 and 5 metres, and uncorrelated normally distributed noise for the orientation in the BEV plane with standard deviation 0.1.

\end{document}